\documentclass[acmsmall,screen]{acmart}
\usepackage{makecell}
\usepackage{multirow}
\usepackage{graphicx}
\usepackage{float}
\usepackage{wrapfig}
\usepackage{tcolorbox} 
\definecolor{myzishe}{HTML}{E6E6FA}
\usepackage{amsmath}
\usepackage{multirow}
\usepackage{multicol}
\usepackage{float}
\usepackage{wrapfig}
\usepackage{tcolorbox} 
\usepackage[table,xcdraw]{xcolor}
\definecolor{myPink}{RGB}{255,226,237}
\definecolor{myTeal}{RGB}{0,161,106}
\definecolor{myOrange}{RGB}{249,70,0}
\definecolor{myOrange1}{RGB}{242,170,132}
\definecolor{nature}{RGB}{101, 164, 135}
\definecolor{medical}{RGB}{104, 52, 154}
\definecolor{myBule}{RGB}{0,104,211}
\definecolor{reda}{RGB}{202,0,0}
\definecolor{mygreen}{RGB}{0,136,51}
\definecolor{myOrange2}{RGB}{197, 90, 17}
\definecolor{myzishe}{HTML}{E6E6FA}
\definecolor{Light}{rgb}{0.92, 0.99, 0.95}
\AtBeginDocument{%
  }

\setcopyright{none}
\renewcommand{\acmVolume}{}
\renewcommand{\acmNumber}{}
\renewcommand{\acmArticle}{}
\renewcommand{\acmMonth}{}
\renewcommand{\acmYear}{}

\begin{document}

\title{TDFNet: Tri-projection Deformable Fusion Network for Panoramic Salient Object Detection}

\author{Qiangqiang Zhou}
\affiliation{
  \institution{School of Artificial Intelligence, Jiangxi Normal University, Nanchang}
  \country{China}
}

\author{iacong Yu}
\authornote{Corresponding author: jiacong\_yu@jxnu.edu.cn}
\affiliation{
  \institution{School of Artificial Intelligence, Jiangxi Normal University, Nanchang}
  \country{China}
}

\author{Jiawei Xu}
\authornote{Corresponding author: Jiawei\_Xu@jxnu.edu.cn.}
\affiliation{
  \institution{School of Artificial Intelligence, Jiangxi Normal University, Nanchang}
  \country{China}
}

\author{Yong Chen}
\affiliation{
  \institution{School of Artificial Intelligence, Jiangxi Normal University, Nanchang}
  \country{China}
}

\author{Xin Huang}
\affiliation{
  \institution{School of Artificial Intelligence, Jiangxi Normal University, Nanchang}
  \country{China}
}

\author{Ping Li}
\affiliation{
  \institution{School of Artificial Intelligence, Jiangxi Normal University, Nanchang}
  \country{China}
}

\renewcommand{\shortauthors}{Trovato et al.}

\begin{abstract}

Recent years have witnessed the growing potential of panoramic salient object detection (PSOD) in robotic vision, virtual reality, and related applications. However, projecting spherical scenes onto 2D planes inevitably introduces geometric distortions, which fundamentally limit the effectiveness of existing projection-based methods. Specifically, Equirectangular Projection (ERP) suffers from severe polar stretching distortions, while cube map projection (CMP) introduces discontinuities across cube-face boundaries, resulting in degraded feature discriminability and compromised geometric consistency. 
To address these limitations, we propose \textbf{TDFNet}, the first \underline{\textbf{T}}ri-projection \underline{\textbf{D}}eformable \underline{\textbf{F}}usion \underline{\textbf{Net}}work for panoramic salient object detection, exploiting complementary projection representations to alleviate geometric distortions and improve detection performance.
Specifically, we design a cross-projection deformable attention (CDA) module that leverages spatial correspondences between different projections to construct geometry-aware sampling locations, guiding deformable attention for cross-projection contextual aggregation and enhancing robustness against projection-induced deformations. Furthermore, we introduce a latitude-guided fusion (LGF) module, which utilizes spherical latitude priors to construct geometric confidence weights for adaptively balancing ERP and CMP features. Meanwhile, LGF incorporates distortion-reduced semantic references from Tangent Projection to achieve cross-projection feature refinement and spatial alignment.
By constructing a three-branch encoding architecture based on ERP, CMP, and Tangent Projection, TDFNet simultaneously preserves global spatial continuity, local geometric details, and fine-grained boundary information. 
Without bells and whistles, TDFNet consistently outperforms existing SOTA methods on four PSOD benchmarks, demonstrating the effectiveness of multi-projection collaborative modeling through both quantitative and qualitative evaluations. We hope that multi-projection collaborative modeling will further advance the development of the panoramic image field. Our code will be made publicly available.
\end{abstract}



\maketitle

\section{Introduction}
With the rapid development of 360-degree imaging technologies~\cite{ni2025makes,yu2023applications,bao2025free360}, Panoramic Salient Object Detection (PSOD) has attracted increasing attention in robotic vision, virtual reality, and other omnidirectional perception applications~\cite{wang2020attention,gungordu2024saliency,yan2025omnidirectional,xiao2026layer}. Unlike conventional perspective images, panoramic images provide a complete field of view, enabling intelligent systems to perceive holistic scene information. However, the spherical nature of panoramic images introduces unique geometric challenges when projected onto 2D representations, resulting in inevitable distortions that hinder accurate feature representation and salient object localization.

Despite substantial progress in PSOD, existing methods~\cite{huang2023lightweight,zhao2023distortion,zhang2023salient,xiao2026staying} remain fundamentally constrained by projection-dependent limitations. Current approaches can be broadly categorized into two groups: ERP-based methods~\cite{huang2023lightweight,long2025enhancing,zhao2023distortion} that directly process Equirectangular Projection (ERP) representations, and dual-projection methods~\cite{zhang2023salient,zhang2022channel,wu2025dagait} that combine ERP with cube map projection (CMP). Although these projection representations provide complementary advantages, neither can fully resolve the geometric inconsistencies introduced during spherical-to-planar mapping. Specifically, ERP preserves global spherical continuity but suffers from severe polar stretching distortion, whereas CMP alleviates polar deformation through local perspective decomposition but introduces semantic discontinuities across cube-face boundaries.

\begin{figure}[t]
\centering
\includegraphics[width=\linewidth]{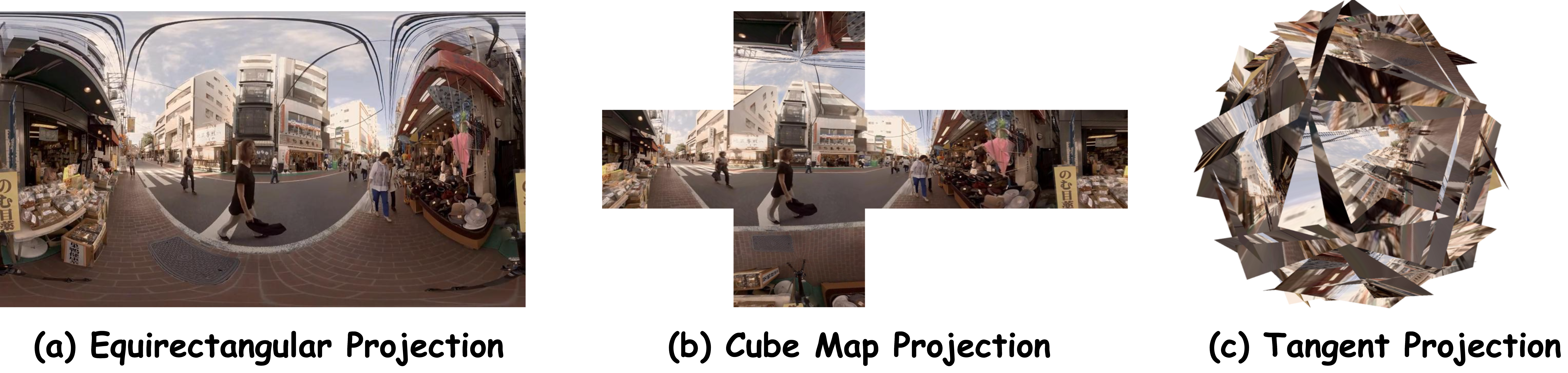}
\caption{Comparison of projection representations used in PSOD. (a) ERP maintains complete spherical topology but suffers from polar stretching distortion. (b) CMP reduces polar deformation through hexahedral decomposition but introduces semantic discontinuities at cube-face boundaries. (c) Tangent Projection, introduced in this work, provides distortion-reduced local perspective views through discrete viewport mapping, complementing ERP and CMP with high-fidelity geometric details.}
\Description{projection}
\label{fig:image1}
\end{figure}

To further analyze these limitations, ERP-only methods have been developed to preserve complete panoramic information within a unified representation. However, the non-uniform sampling distribution caused by ERP projection significantly degrades feature quality in high-latitude regions. For example, the view-aware model proposed in \cite{wu2022view} employs perspective transformations with adaptive fusion to alleviate distortion and scale variations, while DATFormer \cite{zhao2023distortion} introduces distortion maps, adaptive attention, and relational matrices to encode projection-aware spatial priors. Nevertheless, these approaches still struggle to compensate for the intrinsic geometric discrepancy between equatorial and polar regions.

To overcome the limitations of ERP, dual-projection approaches exploit the complementary properties of ERP and CMP. Although these methods improve distortion handling to some extent, the cube-face discontinuities introduced by CMP remain problematic for maintaining object-level structural consistency. For instance, CSMANet \cite{zhang2022channel} adopts a dual-branch architecture with channel-spatial mutual attention to fuse ERP and CMP features, while HPNet \cite{zhang2023salient} develops a hybrid projection feature fusion strategy. However, residual boundary artifacts between cube faces continue to affect coherent object representation in complex panoramic scenes.

To further illustrate the geometric limitations of ERP and CMP, Fig.~\ref{fig:image1} presents three representative panoramic image projections. 
These observations indicate that existing ERP and CMP based approaches still lack a unified projection representation capable of simultaneously preserving global spherical continuity, local geometric fidelity, and boundary consistency. Therefore, exploring additional projection representations and geometry-aware fusion strategies is essential for advancing PSOD.

To address these challenges, we propose TDFNet, the first framework that introduces tri-projection deformable fusion for PSOD. TDFNet constructs a three-branch encoding architecture based on ERP, CMP, and Tangent Projection, where each projection provides complementary geometric information. Specifically, we introduce a cross-projection deformable attention (CDA) module that exploits spatial correspondences between ERP and CMP to generate geometry-aware sampling locations, enabling deformable attention to adaptively aggregate cross-projection contextual information and alleviate projection-induced distortions. Furthermore, we design a latitude-guided fusion (LGF) module that incorporates spherical latitude priors for geometry-guided adaptive fusion between ERP and CMP features, while leveraging distortion-reduced tangent features as semantic references for cross-projection refinement. Through these designs, TDFNet effectively integrates global structural continuity, local geometric fidelity, and fine-grained boundary information for PSOD.

\textbf{Our main contributions are summarized as follows}:
\begin{itemize}
\item We propose TDFNet, the first framework that introduces tri-projection deformable fusion into PSOD. By integrating ERP, CMP, and Tangent Projection, TDFNet captures complementary global structures and low-distortion local geometric details.
\item We develop CDA, a geometry-aware deformable attention module that exploits cross-projection spatial correspondences to alleviate distortions in ERP and CMP representations while improving feature consistency.
\item We design LGF, a geometry-guided adaptive fusion module that utilizes spherical latitude priors and tangent semantic references to achieve efficient alignment and fusion among heterogeneous projection features.
\item Extensive experiments on four widely used PSOD benchmarks demonstrate the effectiveness of TDFNet, achieving consistent improvements over existing state-of-the-art methods in both quantitative and qualitative evaluations.
\end{itemize}
\section{Related Work}
\subsection{Salient object detection in 2D images}
Early salient object detection methods~\cite{jiang2013salient,long2026towards,mazumdar2019content,lebreton2018gbvs360} relied on hand-crafted features and traditional machine learning. With the advent of deep learning, CNNs and vision Transformers have dominated this field by capturing long-range dependencies and global context. For instance, SENet \cite{hao2025simple} introduced a Transformer-based network with an asymmetric encoder-decoder architecture and dynamic weighted loss for enhanced detection. These RGB-based methods achieve strong performance on planar images by exploiting spatial hierarchies and semantic representations; however, their design assumptions are inherently tied to conventional perspective projection, where geometric consistency and uniform spatial distribution are preserved across the image plane.

To overcome unimodal limitations, multimodal methods~\cite{xu2025semantic,tang2024divide,long2025revisiting,wei2025rdfcnet,zhou2026differseg} integrate complementary cues from depth or thermal infrared sensors. For instance, SOMA-Net \cite{xu2025semantic} utilizes sparse semantic enhancement and orthogonal multimodal attention fusion for RGB-D detection. CONTRINET \cite{tang2024divide} employs modality-specific complementary streams for RGB-T detection. Additionally, some studies \cite{xu2026hvpnet,xu2026tp} have developed the biologically inspired HVPNet~\cite{xu2026hvpnet} and  TP-Seg~\cite{xu2026tp}, demonstrating strong potential for lightweight and unified multimodal modeling. Despite these advancements, such 2D methods face fundamental challenges when applied to PSOD. The 360° field of view introduces wraparound continuity alongside severe ERP polar distortion and CMP boundary discontinuities, which exceed the modeling capacity of conventional 2D architectures. Furthermore, multimodal approaches rely on auxiliary data that are scarce in panoramic scenarios, and they remain inherently incapable of resolving projection-induced deformations or viewport-dependent scale variations. Consequently, a dedicated multi-projection architecture remains essential to simultaneously preserve distortion-free local fidelity and maintain global structural continuity.

\subsection{Panoramic Image Saliency Prediction}
Panoramic image saliency prediction estimates continuous fixation density maps to model visual attention in 360° scenes~\cite{zou2023360,lyu2025hallu_vdc,zhu2025scandtm,zhang2024multi}. This task differs from panoramic salient object detection, which requires accurate binary masks and well-defined object boundaries. Early approaches~\cite{yamanaka2023multi,xu2021saliency,zhong2025ctd,zhong2025semi,chao2020multi} adapted conventional 2D models to spherical imagery to alleviate projection distortions. For example, a multiscale framework~\cite{yamanaka2023multi} extracts overlapping 2D patches from multiple viewpoints and introduces learnable equatorial bias layers to capture panoramic attention patterns. EPSNet~\cite{zou2023360} employs a proxy task that localizes cubemap faces within equirectangular projections, enabling joint learning of local and global representations. A multiscale graph network~\cite{zhang2024multi} further constructs superpixel-based graphs with spherical sampling nodes and extracts hierarchical features through graph convolutions and 1D autoencoders before fusion with a U-Net~\cite{ronneberger2015u}. Despite these advances, these methods are optimized for continuous fixation prediction rather than object-level segmentation. Without explicit boundary supervision and binary mask prediction, they cannot provide the precise foreground-background separation required for salient object detection. Therefore, dedicated panoramic salient object detection architectures are still needed to jointly address projection distortion and object-level segmentation.

\begin{figure}[t]
    \centering
    \includegraphics[width=\linewidth]{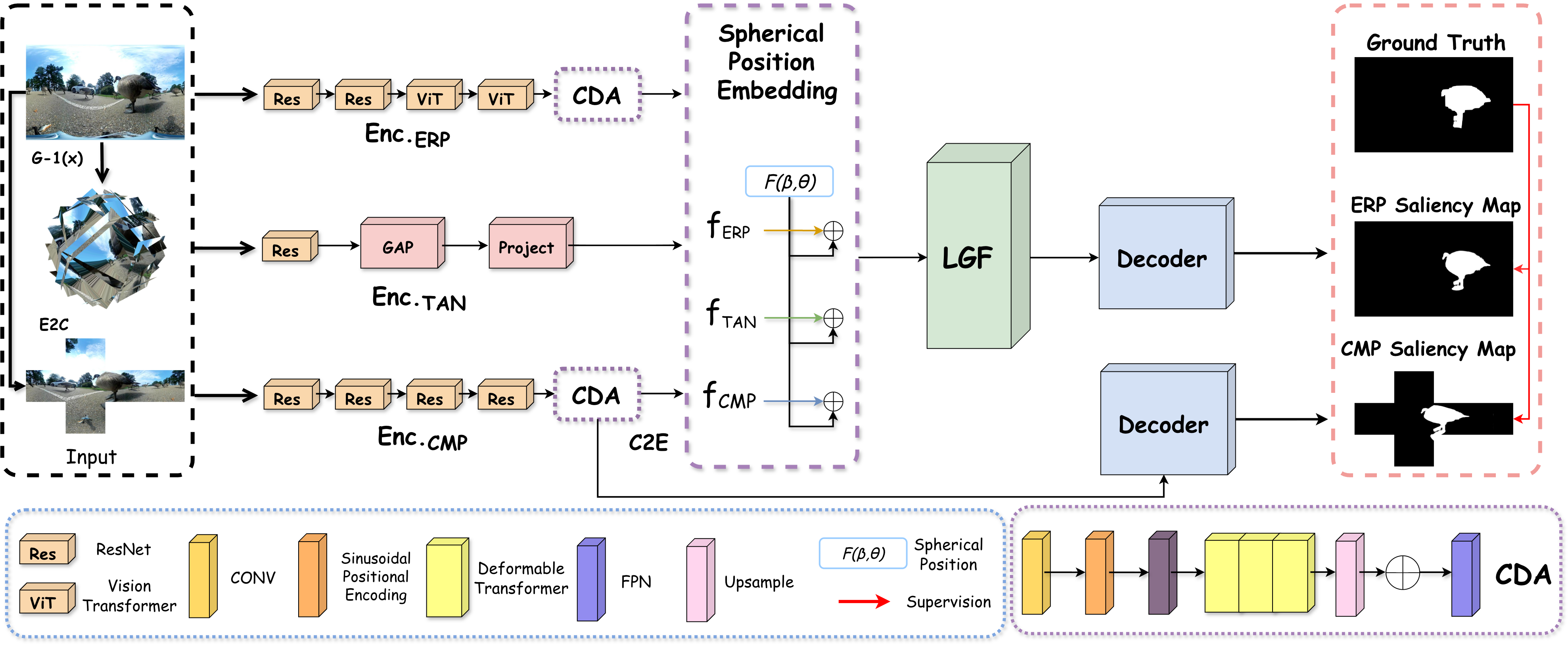}
    \caption{Overall architecture of the proposed TDFNet. The network incorporates three core components: (i) a multi-view tangent projection branch for distortion-free local feature extraction, (ii) the cross-projection deformable attention (CDA) deployed in both ERP and cubemap branches to enhance robustness against projection distortions, and (iii) the latitude-guided fusion (LGF) module that progressively integrates tri-branch features under spherical geometric guidance for cross-projection alignment.}
    \Description{our method}
    \label{fig:image2}
\end{figure}
\subsection{Projection Representations in Panoramic Salient Object Detection}
As discussed above, projection representation is critical for panoramic salient object detection~\cite{zou2023360,zhang2024multi,zhu2025scandtm}, as different formats involve distinct trade-offs between global continuity and local geometric distortion. Equirectangular projection (ERP) provides complete spherical coverage but suffers from severe stretching near the poles. To alleviate this issue, DDS~\cite{li2019distortion} partitions ERP images into blocks and applies adaptive convolutional kernels with multiscale contextual supervision. Similarly, MIDP-Net~\cite{dai2023360} integrates multi-receptive-field features and employs a two-stage decoder combining edge and saliency cues to improve boundary prediction under distortion. In contrast, cubemap projection (CMP) reduces local geometric distortion by decomposing the sphere into six perspective faces, but introduces discontinuities across face boundaries. To exploit their complementary properties, several methods~\cite{huang2020fanet,zhong2024causal,xu2026psgait,he2024sihenet,lyu2026hallu_sae} jointly model ERP and CMP features. For example, SCFANet~\cite{he2023scfanet} uses Vision Transformers for global ERP semantics and CNNs for local CMP details, followed by hierarchical feature interaction. Despite these advances, existing methods~\cite{zhang2023salient,qing2022attentive,zhong2025adaptive,zhang2023360} still mainly rely on ERP and CMP, while tangent-plane projections that provide locally undistorted perspective views remain underexplored. Moreover, existing fusion architectures often introduce redundant parameters, limiting their efficiency in resource-constrained scenarios.
\section{Methods}
As shown in Fig.~\ref{fig:image2}, TDFNet adopts a tri-branch architecture to jointly exploit complementary representations of panoramic images. Given an input ERP image, the ERP branch directly extracts globally continuous features, while the cubemap branch first converts the panorama into six perspective faces to capture locally less-distorted representations. In parallel, we introduce a multi-view tangent projection branch, which samples multiple distortion-free local viewports from the spherical surface to provide fine-grained geometric and boundary cues. To further alleviate projection-induced distortions, Cross-projection deformable attention (CDA) is embedded into both the ERP and cubemap branches, where cross-projection geometric correspondences are used to guide adaptive feature sampling and aggregation. The enhanced ERP and cubemap features are then progressively integrated with tangent features through the latitude-guided fusion (LGF) module. LGF first performs geometry-aware fusion between ERP and cubemap representations according to spherical latitude priors, and subsequently employs tangent features to refine the fused representation across projection domains. Finally, the resulting unified features are fed into the decoder to generate the saliency prediction.

\subsection{Tri‑branch Encoder}

Given an input panoramic image $I_{\mathrm{erp}}\in\mathbb{R}^{3\times H\times W}$, our encoder constructs three complementary representations through ERP, cubemap (CMP), and tangent projection branches. The overall encoding process can be formulated as:
\begin{equation}
\left\{
\begin{aligned}
f_{\mathrm{erp}} &= \mathcal{E}_{\mathrm{erp}}(I_{\mathrm{erp}}),\\
f_{\mathrm{cmp}} &= \mathcal{E}_{\mathrm{cmp}}\big(\mathcal{P}_{\mathrm{E2C}}(I_{\mathrm{erp}})\big),\\
f_{\mathrm{tan}} &= \mathcal{E}_{\mathrm{tan}}\big(\mathcal{P}_{\mathrm{tan}}(I_{\mathrm{erp}})\big),
\end{aligned}
\right.
\label{eq:tribranch}
\end{equation}
where $\mathcal{E}_{\mathrm{erp}}$, $\mathcal{E}_{\mathrm{cmp}}$, and $\mathcal{E}_{\mathrm{tan}}$ denote the corresponding feature encoders, while $\mathcal{P}_{\mathrm{E2C}}$ and $\mathcal{P}_{\mathrm{tan}}$ denote ERP-to-cubemap and tangent projection, respectively. The resulting $F_{\mathrm{erp}}$, $F_{\mathrm{cmp}}$, and $F_{\mathrm{tan}}$ provide globally continuous, locally geometry-preserving, and locally distortion-free representations, respectively.

Specifically, the ERP branch directly feeds $I_{\mathrm{erp}}$ into a Hybrid-ViT-based encoder~\cite{zhang2022channel} built upon ResNet and extracts hierarchical multi-scale features:
\begin{equation}
f_{\mathrm{erp}}
=
\left\{
f_{\mathrm{erp}}^{l}
\right\}_{l=1}^{L}
=
\mathcal{E}_{\mathrm{erp}}(I_{\mathrm{erp}}),
\end{equation}
where $f_{\mathrm{erp}}^{l}\in\mathbb{R}^{C_l\times H_l\times W_l}$ denotes the feature at the $l$-th encoder level. Benefiting from the continuous and regular latitude-longitude sampling grid, the ERP branch preserves the global spatial structure and long-range scene continuity, but inevitably suffers from increasing geometric stretching toward the polar regions.

In parallel, the CMP branch first transforms the ERP panorama into six perspective cube faces through the E2C projection:
\begin{equation}
\left\{
I_{\mathrm{cmp}}^{k}
\right\}_{k=1}^{6}
=
\mathcal{P}_{\mathrm{E2C}}(I_{\mathrm{erp}}),
\end{equation}
where $k$ indexes the six cube faces. Each face is independently encoded using a ResNet-based backbone~\cite{gao2019res2net},
\begin{equation}
\left\{
f_{\mathrm{cmp},k}^{l}
\right\}_{l=1}^{L}
=
\mathcal{E}_{\mathrm{cmp}}
\left(
I_{\mathrm{cmp}}^{k}
\right),
\qquad k=1,\ldots,6.
\end{equation}
The resulting cube-face features are subsequently mapped back to the ERP coordinate domain through the C2E transformation to establish spatial correspondence with the ERP representation:
\begin{equation}
f_{\mathrm{cmp}}^{l}
=
\mathcal{P}_{\mathrm{C2E}}
\left(
\left\{
f_{\mathrm{cmp},k}^{l}
\right\}_{k=1}^{6}
\right),
\end{equation}
yielding the aligned multi-scale representation
\begin{equation}
f_{\mathrm{cmp}}
=
\left\{
f_{\mathrm{cmp}}^{l}
\right\}_{l=1}^{L}.
\end{equation}
Compared with ERP, the local perspective geometry of cubemap faces substantially alleviates polar stretching, although semantic discontinuities may arise around cube-face boundaries.

To further complement these two representations, we introduce Tangent Projection into PSOD as the third feature extraction branch. Unlike its previous use in panoramic video gaze prediction~\cite{cokelek2025spherical}, which mainly performs continuous density regression with relatively relaxed boundary constraints, PSOD requires pixel-level accurate object masks and therefore imposes substantially stronger requirements on spatial correspondence and boundary preservation. Tangent Projection maps local spherical regions onto perspective tangent planes, providing geometrically faithful local observations without ERP polar stretching or CMP face partitioning.

Specifically, we distribute tangent points over four latitude rows containing $3$, $6$, $6$, and $3$ longitudinal centers, respectively, producing $N_t=18$ local tangent viewports. For the $i$-th viewport, each normalized tangent-plane position $(x,y)$ is first mapped back onto the unit sphere through inverse gnomonic projection:
\begin{equation}
\left\{
\begin{aligned}
P_{\mathrm{sphere}}^{(i)}
&=
\mathcal{G}^{-1}
\big(
(x,y);
\theta_i,\phi_i,\mathrm{FOV}
\big),\\
(u,v)
&=
\Psi
\left(
P_{\mathrm{sphere}}^{(i)}
\right)
=
\left(
\frac{\theta}{\pi},
\frac{\phi}{\pi/2}
\right),\\
I_{\mathrm{tan}}^{(i)}
&=
S_{\mathrm{bilinear}}
\left(
I_{\mathrm{erp}},(u,v)
\right),
\end{aligned}
\right.
\label{eq:tangent_projection}
\end{equation}
where $(\theta_i,\phi_i)$ denotes the central longitude and latitude of the $i$-th tangent viewport, $\mathrm{FOV}=80^{\circ}$ denotes its field of view, and $I_{\mathrm{tan}}^{(i)}\in\mathbb{R}^{3\times224\times224}$ is the resulting tangent image. $\mathcal{G}^{-1}$ denotes inverse gnomonic projection, $\Psi(\cdot)$ converts spherical coordinates into the normalized ERP sampling domain, and $S_{\mathrm{bilinear}}$ denotes bilinear grid sampling.

All tangent views are subsequently processed by a frozen ResNet18 backbone. After global average pooling, each viewport produces a $256$-dimensional local representation:
\begin{equation}
f_{\mathrm{tan}}^{(i)}
=
\operatorname{GAP}
\left(
\mathcal{E}_{\mathrm{tan}}
\left(
I_{\mathrm{tan}}^{(i)}
\right)
\right)
\in\mathbb{R}^{256}.
\end{equation}
The complete tangent representation is therefore expressed as
\begin{equation}
f_{\mathrm{tan}}
=
\left[
f_{\mathrm{tan}}^{(1)},
f_{\mathrm{tan}}^{(2)},
\ldots,
f_{\mathrm{tan}}^{(N_t)}
\right]
\in
\mathbb{R}^{N_t\times256},
\qquad N_t=18.
\end{equation}

Accordingly, the tri-branch encoder finally produces
\begin{equation}
\left\{
f_{\mathrm{erp}},
f_{\mathrm{cmp}},
f_{\mathrm{tan}}
\right\}.
\end{equation}
where the three representations respectively emphasize global spatial continuity, locally rectified spherical geometry, and distortion-free fine-grained local semantics. The ERP and CMP representations are subsequently enhanced by CDA, while all three representations are progressively aligned and integrated by LGF.
\subsection{Cross-projection Deformable Attention}
\label{sec:3.2}

Although ERP and CMP provide complementary geometric representations, both still suffer from projection-specific distortions. We therefore propose cross-projection deformable attention (CDA) to enhance the two branches by introducing cross-projection geometric priors into adaptive feature sampling. Given the multi-scale representations from the tri-branch encoder,
$f_{\mathrm{erp}}=\{f_{\mathrm{erp}}^{l}\}_{l=1}^{L}$ and
$f_{\mathrm{cmp}}=\{f_{\mathrm{cmp}}^{l}\}_{l=1}^{L}$,
CDA independently enhances the ERP and CMP features using a shared hybrid reference-point prior:
\begin{equation}
\left\{
\begin{aligned}
\hat{f}_{\mathrm{erp}}
&=
\mathrm{CDA}
\left(
f_{\mathrm{erp}},R
\right),\\
\hat{f}_{\mathrm{cmp}}
&=
\mathrm{CDA}
\left(
f_{\mathrm{cmp}},R
\right),
\end{aligned}
\right.
\label{eq:cda_overall}
\end{equation}
where $R$ denotes the hybrid geometric reference points constructed from the spatial correspondence between ERP and CMP. The resulting
$\hat{f}_{\mathrm{erp}}$ and
$\hat{f}_{\mathrm{cmp}}$ preserve their projection-specific characteristics while improving local geometric consistency.

CDA is applied only to the ERP and CMP branches. The tangent projection branch consists of discrete perspective viewports and does not share the unified latitude--longitude coordinate system used to establish ERP--CMP correspondences. Directly imposing the same reference-point prior on tangent features would therefore introduce coordinate misalignment and undermine their distortion-free local geometry. CDA consists of two steps: Hybrid Reference Point Generation (HRPG) and Deformable Transformer Aggregation.

\begin{figure}[t]
\centering
\includegraphics[width=\linewidth]{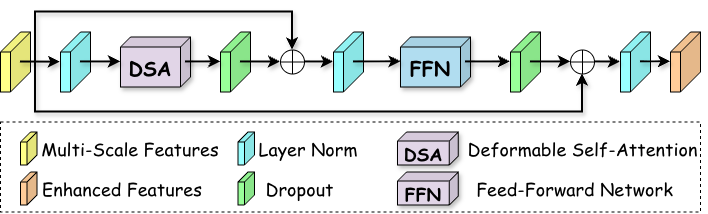}
\caption{The structure of the deformable Transformer block in the CDA module.}
\Description{Deformable Transformer block in CDA.}
\label{fig:image3}
\end{figure}
\textbf{Hybrid Reference Point Generation.}
For each branch $b\in\{\mathrm{erp},\mathrm{cmp}\}$, CDA takes the hierarchical features
$f_{b}=\{f_b^{l}\}_{l=1}^{L}$ as input. The latter three feature levels are first projected to a common channel dimension $d=256$ using $1\times1$ convolutions:
\begin{equation}
X_b^{l}
=
\mathrm{Conv}_{1\times1}
\left(
F_b^{l}
\right),
\qquad
X_b^{l}\in
\mathbb{R}^{d\times H_l\times W_l}.
\end{equation}

To retain explicit spatial information, a two-dimensional sinusoidal positional encoding $P_l$ is added to each projected feature:
\begin{equation}
\tilde{X}_b^{l}
=
X_b^{l}+P_l.
\end{equation}
For a spatial location $(i,j)$ with normalized coordinates
$x=j/W_l$ and $y=i/H_l$, the positional encoding is defined as
\begin{equation}
\left\{
\begin{aligned}
P_l(i,j)_{2k}
&=
\sin\left(
\frac{x}{T^{4k/d}}
\right),
&
P_l(i,j)_{2k+1}
&=
\cos\left(
\frac{x}{T^{4k/d}}
\right),
\\
P_l(i,j)_{2k+\frac{d}{2}}
&=
\sin\left(
\frac{y}{T^{4k/d}}
\right),
&
P_l(i,j)_{2k+1+\frac{d}{2}}
&=
\cos\left(
\frac{y}{T^{4k/d}}
\right),
\end{aligned}
\right.
\label{eq:cda_pe}
\end{equation}
where $k=0,\ldots,d/4-1$ and $T=10000$.

Hybrid Reference Point Generation then constructs geometric reference points for deformable sampling. For the $l$-th feature level, the basic ERP reference point at location $(i,j)$ is defined as
\begin{equation}
\mathbf{r}_{l,ij}^{\mathrm{ERP}}
=
\left(
\frac{j+0.5}{W_l\,r_{l,w}},
\frac{i+0.5}{H_l\,r_{l,h}}
\right),
\label{eq:erp_reference}
\end{equation}
where $r_{l,w}$ and $r_{l,h}$ denote the corresponding spatial scaling ratios.

Based on this anchor, additional cross-projection reference points are distributed along eight directions,
$\theta_m=\pi m/4$, $m=0,\ldots,7$, and four radial levels,
$\rho_n=n/4$, $n=0,\ldots,3$:
\begin{equation}
\mathbf{r}_{l,ij}^{(m,n)}
=
\mathrm{clamp}
\left(
\mathbf{r}_{l,ij}^{\mathrm{ERP}}
+
\rho_n
\begin{bmatrix}
\cos\theta_m\\
\sin\theta_m
\end{bmatrix},
[0,1]^2
\right).
\label{eq:cmp_reference}
\end{equation}

The ERP anchor and its cross-projection neighborhood are then grouped to form the hybrid reference-point set:
\begin{equation}
R
=
\mathrm{Concat}
\left(
\left\{
\mathbf{r}_{l,ij}^{\mathrm{ERP}}
\right\}_{n=0}^{3},
\left\{
\mathbf{r}_{l,ij}^{(m,n)}
\right\}_{m=0,n=0}^{7,3}
\right),
\label{eq:hybrid_reference}
\end{equation}
which provides a geometry-aware sampling prior for both projection branches. In this way, the initial sampling locations are no longer determined solely by the regular feature grid, but are explicitly constrained by cross-projection geometric neighborhoods.

\textbf{Deformable Transformer Aggregation.}
After constructing $R$, the projected multi-scale features are flattened and concatenated into a unified sequence:
\begin{equation}
X_b
=
\mathrm{Concat}_{l}
\left(
\mathrm{Flatten}
\left(
\tilde{X}_b^{l}
\right)
\right).
\end{equation}

As illustrated in Fig.~\ref{fig:image3}, the sequence is processed by multiple deformable transformer blocks. For a query feature $\mathbf{q}$, deformable self-attention predicts a set of sampling offsets $\Delta\mathbf{p}$ and normalized attention weights $\mathbf{A}$:
\begin{equation}
\left\{
\begin{aligned}
\Delta\mathbf{p}
&=
\mathbf{W}_{\mathrm{off}}\mathbf{q},\\
\mathbf{A}
&=
\operatorname{Softmax}
\left(
\mathbf{W}_{\mathrm{attn}}\mathbf{q}
\right),
\end{aligned}
\right.
\label{eq:cda_offset}
\end{equation}
where $\mathbf{W}_{\mathrm{off}}$ and $\mathbf{W}_{\mathrm{attn}}$ are learnable projection matrices.

The reference points are dynamically adjusted by the learned offsets:
\begin{equation}
\hat{\mathbf{p}}
=
\mathbf{R}
+
\Delta\mathbf{p}.
\end{equation}

The corresponding features are then obtained through bilinear sampling and aggregated according to their attention weights:
\begin{equation}
\mathbf{z}
=
\sum_{l=1}^{L}
\sum_{p=1}^{P}
\mathbf{A}_{p}^{(l)}
\,
\mathrm{BilinearSample}
\left(
F_b^{l},
\hat{\mathbf{p}}_{p}^{(l)}
\right),
\label{eq:cda_aggregation}
\end{equation}
where $P$ denotes the number of sampling points and $\mathbf{z}$ is the geometry-enhanced representation for the query.

Each deformable transformer block follows a pre-normalization residual structure:
\begin{equation}
\left\{
\begin{aligned}
Y_b
&=
X_b+
\mathrm{DSA}
\left(
\mathrm{LN}(X_b),R
\right),\\
Z_b
&=
Y_b+
\mathrm{FFN}
\left(
\mathrm{LN}(Y_b)
\right).
\end{aligned}
\right.
\label{eq:cda_block}
\end{equation}

After multiple blocks, the enhanced sequence is reshaped back to its corresponding spatial resolutions and integrated with high-resolution encoder features through an FPN~\cite{lin2017feature}, producing
\begin{equation}
\left\{
\hat{f}_{\mathrm{erp}},
\hat{f}_{\mathrm{cmp}}
\right\}
=
\mathrm{FPN}
\left(
Z_{\mathrm{erp}},
Z_{\mathrm{cmp}}
\right).
\label{eq:cda_output}
\end{equation}

Through the shared hybrid reference-point prior and adaptive deformable sampling, CDA enables each projection branch to exploit geometrically complementary neighborhoods while retaining its own representation characteristics. The enhanced ERP and CMP features,
$\hat{f}_{\mathrm{erp}}$ and
$\hat{f}_{\mathrm{cmp}}$, are subsequently passed to LGF together with the tangent representation $f_{\mathrm{tan}}$ for tri-branch cross-projection fusion.

\subsection{Latitude-Guided Fusion}
\label{sec:3.3}
\begin{wrapfigure}{r}{0.45\textwidth}
    \centering
    \includegraphics[width=0.44\textwidth]{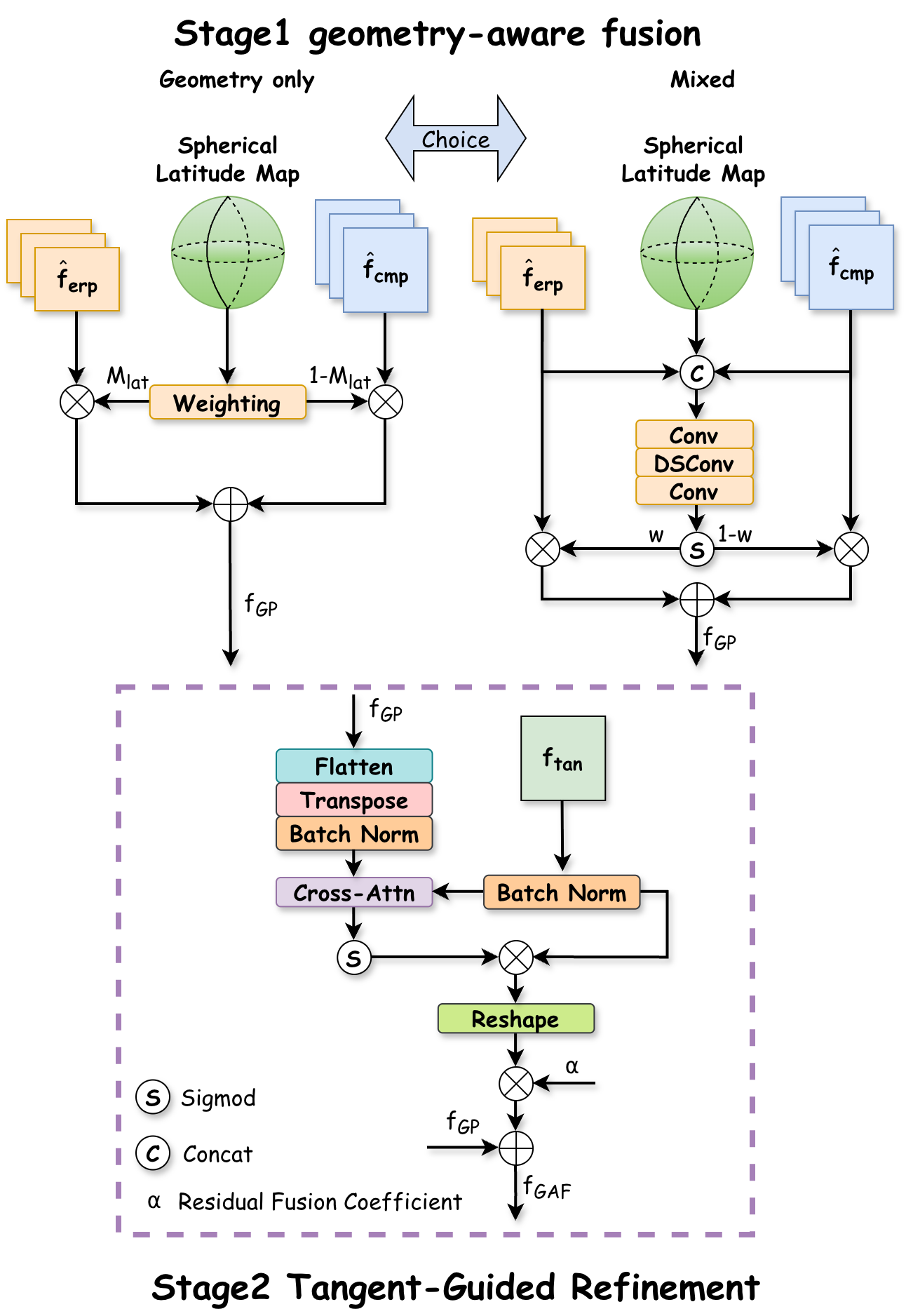}
    \caption{The structure of the LGF module.}
    \Description{LGF module}
    \label{fig:image4}
\end{wrapfigure}
After CDA, the ERP and CMP branches produce distortion-enhanced representations $\hat{f}_{\mathrm{erp}}$ and $\hat{f}_{\mathrm{cmp}}$, while the tangent branch provides a set of locally distortion-free representations $f_{\mathrm{tan}}$. Although these three projections encode complementary information, they exhibit substantially different geometric properties and spatial organizations. ERP maintains a globally continuous latitude--longitude layout and preserves long-range scene structure, but its sampling density becomes increasingly non-uniform toward the poles, leading to severe geometric stretching. CMP mitigates such polar distortion through local perspective projection, yet the partition into six independent faces introduces artificial discontinuities around cube boundaries. In contrast, Tangent Projection preserves local perspective geometry without polar stretching or cube-face truncation, but its discrete viewports form a sparse set of local observations rather than a continuous spherical feature map.
These heterogeneous representations differ in both geometric distortion and spatial organization, making direct fusion prone to cross-projection inconsistency. To address this, LGF first integrates ERP and CMP features using either a parameter-free \textit{Geometry Only} strategy based on spherical latitude or a learnable \textit{Mixed} strategy that combines latitude priors with image content. The fused representation is then refined using tangent features as locally distortion-free semantic references.

To address this issue, as shown in Fig.~\ref{fig:image4}, we propose ratitude-guided fusion (LGF), which progressively integrates the three projection representations through two stages:
\begin{equation}
\left\{
\begin{aligned}
f_{\mathrm{GP}}
&=
\mathcal{G}
\left(
\hat{f}_{\mathrm{erp}},
\hat{f}_{\mathrm{cmp}},
M_{\mathrm{lat}}
\right),\\
f_{\mathrm{GAF}}
&=
\mathcal{T}
\left(
f_{\mathrm{GP}},
f_{\mathrm{tan}}
\right),
\end{aligned}
\right.
\label{eq:lgf_overall}
\end{equation}
where $\hat{f}_{\mathrm{erp}}$ and $\hat{f}_{\mathrm{cmp}}$ denote the spatially aligned bottleneck features obtained from
$\hat{f}_{\mathrm{erp}}$ and
$\hat{f}_{\mathrm{cmp}}$, respectively.
$\mathcal{G}(\cdot)$ denotes geometry-aware ERP--CMP fusion, and
$\mathcal{T}(\cdot)$ denotes Tangent-Guided Refinement.
The output $f_{\mathrm{GAF}}$ is subsequently fed into the decoder for saliency prediction.

We first construct a normalized latitude map according to the spatial resolution of the bottleneck feature:
\begin{equation}
M_{\mathrm{lat}}(i,j)
=
\frac{2i}{H-1}-1,
\qquad
M_{\mathrm{lat}}\in[-1,1]^{H\times W},
\label{eq:lat_map}
\end{equation}
where $H$ and $W$ denote the feature height and width. Thus, $M_{\mathrm{lat}}=0$ corresponds to the equatorial region, whereas $|M_{\mathrm{lat}}|=1$ corresponds to the polar regions. Rather than directly treating latitude as semantic confidence, we regard it as a projection-level geometric prior describing the relative reliability of ERP and CMP under spherical projection. Based on this prior, LGF provides two alternative fusion strategies.

\textit{\textbf{Geometry Only}.}
The Geometry Only strategy directly converts the latitude prior into a deterministic fusion weight without introducing additional learnable parameters:
\begin{equation}
W_{\mathrm{geo}}
=
\cos
\left(
M_{\mathrm{lat}}\frac{\pi}{2}
\right).
\label{eq:geo_weight}
\end{equation}
The corresponding ERP--CMP fusion is formulated as
\begin{equation}
f_{\mathrm{GP}}^{\mathrm{geo}}
=
W_{\mathrm{geo}}
\odot
\hat{f}_{\mathrm{erp}}
+
\left(
1-W_{\mathrm{geo}}
\right)
\odot
\hat{f}_{\mathrm{cmp}},
\label{eq:geo_fusion}
\end{equation}
where $\odot$ denotes element-wise multiplication. Near the equator, $W_{\mathrm{geo}}\rightarrow1$, so ERP contributes more strongly; toward the poles, $W_{\mathrm{geo}}\rightarrow0$, increasing the contribution of CMP features.

\textit{\textbf{Mixed}.}
Although latitude provides a stable geometric prior, geometric reliability does not always coincide with semantic reliability. For example, CMP may be geometrically preferable at high latitudes but can still introduce semantic discontinuities when an object crosses cube-face boundaries. Therefore, the Mixed strategy jointly considers spherical geometry and image content.

The enhanced ERP and CMP features are first concatenated with the latitude map:
\begin{equation}
X_{\mathrm{mix}}
=
\operatorname{Concat}
\left(
\hat{f}_{\mathrm{erp}},
\hat{f}_{\mathrm{cmp}},
M_{\mathrm{lat}}
\right).
\end{equation}
A lightweight convolutional subnetwork then predicts a spatially adaptive fusion weight:
\begin{equation}
W_{\mathrm{mix}}
=
\sigma
\left(
\mathrm{Conv}_{1\times1}^{(2)}
\left(
\mathrm{DSConv}
\left(
\mathrm{Conv}_{1\times1}^{(1)}
\left(
X_{\mathrm{mix}}
\right)
\right)
\right)
\right),
\label{eq:mixed_weight}
\end{equation}
where $\mathrm{DSConv}$ denotes depthwise separable convolution~\cite{howard2017mobilenets}, and $\sigma(\cdot)$ denotes the Sigmoid function. The corresponding fused representation is
\begin{equation}
f_{\mathrm{GP}}^{\mathrm{mix}}
=
W_{\mathrm{mix}}
\odot
\hat{f}_{\mathrm{erp}}
+
\left(
1-W_{\mathrm{mix}}
\right)
\odot
\hat{f}_{\mathrm{cmp}}.
\label{eq:mixed_fusion}
\end{equation}
Unlike Geometry Only, the latitude map here serves as an explicit geometric condition rather than a fixed fusion rule, allowing the fusion weights to adapt to local semantic content.

\textbf{Tangent-Guided Refinement.}
Both fusion strategies subsequently employ the same Tangent-Guided Refinement module. Let $f_{\mathrm{tan}}$
denote the tangent-view representations. For either
$s\in\{\mathrm{geo},\mathrm{mix}\}$,
the fused ERP--CMP representation is flattened as:
\begin{equation}
X_{\mathrm{GP}}^{s}
=
\operatorname{Flatten}
\left(
f_{\mathrm{GP}}^{s}
\right)
\in
\mathbb{R}^{N_s\times d},
\qquad
N_s=HW.
\end{equation}

The fused spherical features form the queries, while the tangent-view features provide the keys and values:
\begin{equation}
\left\{
\begin{aligned}
Q^{s}
&=
W_Q
\operatorname{BN}
\left(
X_{\mathrm{GP}}^{s}
\right),\\
K
&=
W_K
\operatorname{BN}
\left(
\mathcal{F}_{\mathrm{tan}}
\right),\\
V
&=
W_V
\operatorname{BN}
\left(
\mathcal{F}_{\mathrm{tan}}
\right).
\end{aligned}
\right.
\label{eq:lgf_qkv}
\end{equation}
The cross-projection attention is computed as
\begin{equation}
A^{s}
=
\operatorname{Softmax}
\left(
\beta Q^{s}K^{\top}
\right)
\in
\mathbb{R}^{N_s\times N_t},
\label{eq:lgf_attention}
\end{equation}
where $\beta$ is empirically set to $10$. The tangent-guided refinement feature is then obtained by
\begin{equation}
f_{\mathrm{r}}^{s}
=
\operatorname{Reshape}
\left(
A^{s}V
\right)
\in
\mathbb{R}^{d\times H\times W}.
\label{eq:tangent_refinement}
\end{equation}
In this way, each spatial location selectively retrieves complementary local semantics from the $18$ distortion-free tangent viewpoints.

Finally, the two strategies employ different residual fusion coefficients:
\begin{equation}
f_{\mathrm{GAF}}^{s}
=
f_{\mathrm{GP}}^{s}
+
\alpha_{s}
f_{\mathrm{r}}^{s},
\qquad
s\in\{\mathrm{geo},\mathrm{mix}\},
\label{eq:lgf_output}
\end{equation}
where
\begin{equation}
\alpha_{s}
=
\begin{cases}
0.15, & s=\mathrm{geo},\\[2mm]
0.5\,\sigma(\hat{\alpha}), & s=\mathrm{mix},
\end{cases}
\label{eq:alpha_strategy}
\end{equation}
and $\hat{\alpha}$ is a learnable scalar. Geometry Only therefore maintains a fixed and geometry-dominated refinement strength, whereas Mixed allows the contribution of tangent semantics to adapt during training.

\subsection{Decoder and Loss Function}

The fused bottleneck features from LGF are fed into the ERP decoder to generate the final saliency prediction map $P_{\mathrm{final}}$. 
Meanwhile, a cube auxiliary decoder is introduced to provide additional supervision in the cubemap domain. 
The auxiliary prediction $P_{\mathrm{aux}}$ is only used during training and is discarded during inference. 
Both decoders adopt a multi-scale refinement structure to progressively integrate encoder features and recover high-resolution saliency maps.

TDFNet is optimized with the structure loss~\cite{li2021salient}, which combines weighted binary cross-entropy and weighted IoU:
\begin{equation}
\mathcal{L}_{\mathrm{structure}}(S,G)
=
\mathcal{L}_{\mathrm{BCE}}^{w}(S,G)
+
\mathcal{L}_{\mathrm{IoU}}^{w}(S,G),
\end{equation}
where $S$ and $G$ denote the prediction and ground truth, respectively. 
$\mathcal{L}_{\mathrm{BCE}}^{w}$ and $\mathcal{L}_{\mathrm{IoU}}^{w}$ represent the weighted binary cross-entropy and weighted IoU losses, and $w$ denotes the pixel-wise weight map. 
This loss jointly optimizes pixel-level accuracy and structural consistency.

During training, both ERP and Cube branches are supervised:
\begin{equation}
\mathcal{L}_{\mathrm{total}}
=
\mathcal{L}_{\mathrm{ERP}}
+
\mathcal{L}_{\mathrm{Cube}},
\end{equation}
where
$\mathcal{L}_{\mathrm{ERP}}
=
\mathcal{L}_{\mathrm{structure}}(P_{\mathrm{final}},G_{\mathrm{ERP}})$
and
$\mathcal{L}_{\mathrm{Cube}}
=
\mathcal{L}_{\mathrm{structure}}(P_{\mathrm{aux}},G_{\mathrm{Cube}})$.
\section{EXPERIMENTS}
\subsection{Dataset}
We evaluate TDFNet on four benchmark datasets for 360° panoramic image salient object detection: 360-SOD\cite{li2019distortion}, 360-SSOD\cite{ma2020stage}, F-360iSOD\cite{zhang2020fixation}, and ODI-SOD\cite{wu2022view}. As the first dataset in this domain, 360-SOD contains 500 high-resolution panoramic images with pixel-level annotations. 360-SSOD extends the scale to 1,105 images, covering 10 semantic categories with more balanced object distribution. F-360iSOD is the first instance-level annotated dataset, providing dense annotations of 1,165 salient objects across 72 categories within 107 images. ODI-SOD represents the largest-scale benchmark to date, comprising 6,263 panoramic images with resolutions no lower than 2K. To ensure fair comparison, we strictly follow the original data splits for all four datasets in our experiments.
\subsection{Implementation Details}
We conduct all experiments using a single NVIDIA RTX 3090 GPU under the PyTorch framework. During training, only ERP-format images are used as input, with spatial resolution uniformly resized to $640 \times 1280$, and no data augmentation strategies are employed. The model is trained end-to-end using the Adam optimizer \cite{kingma2014adam} with a batch size of 2, an initial learning rate 
of $1 \times 10^{-4}$, and a weight decay of $1 \times 10^{-4}$ for 50 epochs. The learning rate schedule combines linear warmup with cosine annealing, accompanied by gradient clipping with a threshold of 0.5 to stabilize the optimization process.

\subsection{Evaluation Metrics}
To comprehensively evaluate the quality of saliency maps generated by the proposed model, we adopt four standard evaluation metrics widely recognized in the field of salient object detection: S-measure ($S_m$)\cite{fan2017structure}, which assesses the structural similarity between the predicted map and the ground truth; mean absolute error (MAE)~\cite{perazzi2012saliency}, which measures the pixel-wise prediction deviation; F-measure ($F_\beta$)~\cite{achanta2009frequency}, which jointly considers the balance between precision and recall; and E-measure ($E_m$)~\cite{fan2018enhanced}, which evaluates the accuracy of salient object detection by jointly capturing image-level and pixel-level statistics.

\subsection{Comparisons with SOTAs}
To comprehensively evaluate the performance of the proposed method, we compare TDFNet against 19 state-of-the-art salient object detection methods. Among them, 9 are SOD methods designed for \(360^{\circ}\) panoramic images, including SIHENet~\cite{he2024sihenet}, MPFR~\cite{cong2023multi}, SCFA~\cite{he2023scfanet}, View~\cite{wu2022view}, LD~\cite{huang2023lightweight}, DDS~\cite{li2019distortion}, FA~\cite{huang2020fanet}, DT~\cite{zhao2023distortion}, and CPNet~\cite{wen2025consistency}; the remaining 10 are SOD methods for 2D images, including BIPG~\cite{yao2021boundary}, PG~\cite{xie2022pyramid}, GFI~\cite{zhu2021supplement}, SCW~\cite{yu2021structure}, HVP~\cite{liu2020lightweight}, ACCo~\cite{li2022adjacent}, BASNet~\cite{qin2021boundary}, DCNet~\cite{zhu2023dc}, BBRF~\cite{ma2023boosting}, and MDSAM~\cite{gao2024multi}. To ensure fairness in comparison, the saliency maps of all competing methods were either directly provided by the authors or generated using code provided by the authors.
\begin{table}[t]
  \centering
  \caption{Quantitative comparison results on \textbf{360-SOD} dataset. ``Ours'' adopts the \textbf{Geometry Only} strategy in the LGF module. $\uparrow$ and $\downarrow$ indicate that the larger scores and the smaller ones are better, respectively. 
    The best scores are highlighted in \textbf{bold}.}
  \label{tab:table1}
  \small
  \setlength{\tabcolsep}{8pt}
  \begin{tabular}{rcccccc}
    \toprule
    \textbf{Model} & \textbf{$S_m\uparrow$} & \textbf{$MAE\downarrow$} & \textbf{$meE\uparrow$} & \textbf{$maxE\uparrow$} & \textbf{$meF\uparrow$} & \textbf{$maxF\uparrow$} \\
    \midrule
    FA~\cite{huang2020fanet}      & .826 & .021 & .885 & .900 & .748 & .770 \\
    DDS~\cite{li2019distortion}    & .799 & .023 & .865 & .904 & .695 & .722 \\
    DT~\cite{zhao2023distortion}     & .849 & .017 & .907 & .919 & .774 & .793 \\
    GFI~\cite{zhu2021supplement}    & .831 & .021 & .894 & .900 & .753 & .769 \\
    SCW~\cite{yu2021structure}    & .830 & .021 & .886 & .887 & .754 & .759 \\
    BIPG~\cite{yao2021boundary}   & .811 & .024 & .885 & .890 & .727 & .740 \\
    PG~\cite{xie2022pyramid}     & .750 & .030 & .745 & .786 & .629 & .646 \\
    BAS~\cite{qin2021boundary}    & .654 & .110 & .651 & .676 & .436 & .444 \\
    BBRF~\cite{ma2023boosting}   & .730 & .067 & .759 & .762 & .602 & .609 \\
    ACCo~\cite{li2022adjacent}   & .770 & .025 & .768 & .869 & .676 & .735 \\
    DC~\cite{zhu2023dc}     & .682 & .100 & .705 & .737 & .492 & .507 \\
    MDSAM~\cite{gao2024multi}  & .776 & .055 & .794 & .802 & .641 & .661 \\
    LD~\cite{huang2023lightweight}     & .768 & .029 & .873 & .866 & .641 & .656 \\
    MPFR~\cite{cong2023multi}    & .842 & .019 & .875 & .885 & .755 & .765 \\
    SCFA~\cite{he2023scfanet}   & .871 & .018 & .925 & .930 & .808 & .824  \\
    CPNet~\cite{wen2025consistency}  & .862 & .018 & .875 & ---  & .800 & ---  \\
    \rowcolor{myzishe}
    \textbf{Ours}        & \textbf{.883} & \textbf{.015} & \textbf{.927} & \textbf{.932} & \textbf{.835} &\textbf{.847} \\
    \bottomrule
  \end{tabular}
\end{table}

\begin{table}[t]
  \centering
  \caption{Quantitative comparison results on \textbf{360-SSOD} dataset. ``Ours'' adopts the \textbf{Mixed} strategy in the LGF module. $\uparrow$ and $\downarrow$ indicate that the larger scores and the smaller ones are better, respectively.
    The best scores are highlighted in \textbf{bold}.}
  \label{tab:table2}
  \small
  \setlength{\tabcolsep}{8pt}
  \begin{tabular}{rcccccc}
    \toprule
    \textbf{Model} & \textbf{$S_m\uparrow$} & \textbf{$MAE\downarrow$} & \textbf{$meE\uparrow$} & \textbf{$maxE\uparrow$} & \textbf{$meF\uparrow$} & \textbf{$maxF\uparrow$} \\
    \midrule
    
    LD~\cite{huang2023lightweight}     & .756 & .034 & .845 & .863 & .492 & .511 \\
    FA~\cite{huang2020fanet}      & .717 & .039 & .727 & .735 & .520 & .532 \\
    HVP~\cite{liu2020lightweight}    & .773 & .143 & .800 & .854 & .500 & .529 \\
    DT~\cite{zhao2023distortion}     & .770 & .026 & .832 & .865 & .644 & .657 \\
    GFI~\cite{zhu2021supplement}    & .767 & .034 & .829 & .848 & .526 & .536 \\
    SCW~\cite{yu2021structure}    & .760 & .029 & .813 & .854 & .511 & .518 \\
    BIPG~\cite{yao2021boundary}   & .760 & .030 & .817 & .855 & .517 & .524 \\
    PG~\cite{xie2022pyramid}     & .712 & .041 & .731 & .790 & .425 & .452 \\
    BAS~\cite{qin2021boundary}    & .588 & .143 & .604 & .612 & .266 & .271 \\
    ACCo~\cite{li2022adjacent}   & .747 & .031 & .758 & .853 & .465 & .503 \\
    SCFA~\cite{he2023scfanet}   & .791& .039 & .860 &.870 & .560 & .570 \\
    SIHE~\cite{he2024sihenet}    & .788& .028 & .871 &\textbf{.886} & .567 & .576 \\
    CPNet~\cite{wen2025consistency}  & .666 & .052 & .723 & ---  & .474 & ---  \\
    \rowcolor{myzishe}
    \textbf{Ours}        & \textbf{.804}&\textbf{.025} & \textbf{.877} & .883 & \textbf{.697}&\textbf{.708} \\
    \bottomrule
  \end{tabular}
\end{table}

\textbf{Quantitative evaluation:}
We compare TDFNet with existing state-of-the-art methods on four benchmark datasets, including 360-SOD~\cite{li2019distortion}, 360-SSOD~\cite{ma2020stage}, ODI-SOD~\cite{wu2022view}, and F-360iSOD~\cite{zhang2020fixation}.  Overall, TDFNet achieves consistently superior performance across different 360-degree salient object detection benchmarks, demonstrating its effectiveness in handling projection distortion and complex panoramic structures.

\textbf{\textit{360-SOD Dataset}}: In Table~\ref{tab:table1}, we present the quantitative comparison results of the proposed TDFNet against other state-of-the-art methods on the 360-SOD dataset. As can be observed, TDFNet achieves superior performance across the majority of metrics. Compared with the SOTA method SCFA~\cite{he2023scfanet}, our approach achieves superior performance across all six metrics, with S-measure, MAE, mean E-measure, max E-measure, mean F-measure, and max F-measure of 0.883, 0.015, 0.927, 0.932, 0.835, and 0.847, respectively. Specifically, MAE is reduced by 16.7\%, while mean F-measure and max F-measure are improved by 3.3\% and 2.8\%, respectively, demonstrating the overall superiority of TDFNet..

\begin{table}[t]
  \centering
  \caption{Quantitative comparison results on \textbf{ODI-SOD} dataset. ``Ours'' adopts the \textbf{Mixed} strategy in the LGF module. $\uparrow$ and $\downarrow$ indicate that the larger scores and the smaller ones are better, respectively.
    The best scores are highlighted in \textbf{bold}.}
  \label{tab:table3}
  \small
  \setlength{\tabcolsep}{8pt}
  \begin{tabular}{rcccccc}
    \toprule
    \textbf{Model} & \textbf{$S_m\uparrow$} & \textbf{$MAE\downarrow$} & \textbf{$meE\uparrow$} & \textbf{$maxE\uparrow$} & \textbf{$meF\uparrow$} & \textbf{$maxF\uparrow$} \\
    \midrule
    View~\cite{wu2022view}    & .831 & .035 & ---  & ---  & ---  &.822\\
    FA~\cite{huang2020fanet}       & .730 & .050 & .771 & .790 & .610 & .632 \\
    DDS~\cite{li2019distortion}     & .791 & .045 & ---  & ---  & ---  & .761 \\
    HVP~\cite{liu2020lightweight}      & .732 & .061 & .787 & .795 & .616 & .627 \\
    DC~\cite{zhu2023dc}      & .659 & .120 & .685 & .691 & .520 & .530 \\
    MDSAM~\cite{gao2024multi}   & .706 & .084 & .744 & .748 & .597 & .603 \\
    GFI~\cite{zhu2021supplement}     & .779 & .051 & .801 & .810 & .692 & .700 \\
    SCW~\cite{yu2021structure}     & .814 & .043 & .852 & .859 & .745 & .753 \\
    BIPG~\cite{yao2021boundary}    & .815 & .042 & .861 & .867 & .744 & .759 \\
    PG~\cite{xie2022pyramid}      & .808 & .044 & .851 & .857 & .727 & .743 \\
    BBRF~\cite{ma2023boosting}    & .678 & .093 & .717 & .725 & .559 & .561 \\
    ACCo~\cite{li2022adjacent}    & .681 & .094 & .725 & .865 & .551 & .754 \\
    SCFA~\cite{he2023scfanet}    & .840 & .037 & .880& .886& .776& .793 \\
    SIHE~\cite{he2024sihenet}     & .846 & .036 &.888& .894 & .789& .808 \\
    \rowcolor{myzishe}
    \textbf{Ours}         & \textbf{.876} & \textbf{.028} & \textbf{.917} &\textbf{ .923} & \textbf{.833} & \textbf{.845}\\

    \bottomrule
  \end{tabular}
\end{table}

\textbf{\textit{360-SSOD Dataset}}:
In Table~\ref{tab:table2}, we compare the performance of the proposed method with other SOD methods on the 360-SSOD dataset. As shown, our method achieves the best results across most evaluation metrics, including S-measure, MAE, mean F-measure, max F-measure, and mean E-measure. For the max E-measure metric, our method ranks second, with a marginal gap from the top-performing result.

\textbf{\textit{ODI-SOD Dataset}}:
In Table~\ref{tab:table3}, we present a quantitative comparison of the proposed TDFNet with other state-of-the-art SOD methods on the ODI-SOD dataset. As shown, our model achieves optimal performance across all evaluation metrics, attaining 0.876, 0.028, 0.833, 0.845, 0.917, and 0.923 for S-measure, MAE, mean F-measure, max F-measure, mean E-measure, and max E-measure, respectively. Compared to the second-best method, TDFNet reduces MAE by 20.0\% and achieves performance improvements of 3.27\% and 5.58\% in mean E-measure and mean F-measure, respectively.

\begin{table}[t]
  \centering
  \caption{Quantitative comparison results on \textbf{F-360iSOD} dataset. ``Ours'' adopts the \textbf{Geometry Only} strategy in LGF. $\uparrow$ and $\downarrow$ indicate that the larger scores and the smaller ones are better, respectively. The best scores are highlighted in \textbf{bold}.}
  \label{tab:table4}
  \begin{tabular}{rcccccc}
    \toprule
    \textbf{Model} & \textbf{$S_m\uparrow$} & \textbf{$MAE\downarrow$} & \textbf{$meE\uparrow$} & \textbf{$meF\uparrow$} \\
    \midrule
    FA~\cite{huang2020fanet}      & .621& .056 & .700 & .345 \\
    DDS~\cite{li2019distortion}      & .612 & .057 & .700 & .325 \\
    ACCo~\cite{li2022adjacent}    & .613 & .060 & .694 & .334 \\
    MPFR~\cite{cong2023multi}     & .617 & .052&.742 & .370\\
    DT~\cite{zhao2023distortion}      & .611 & .058 & .652 & .310 \\
    CPNet~\cite{wen2025consistency}   & .651 & .051 &.729&.382 \\
    \rowcolor{myzishe}
    \textbf{Ours}         &\textbf{.745} & \textbf{.030} & \textbf{.762} & \textbf{.545} \\
    \bottomrule
  \end{tabular}
\end{table}

\textbf{\textit{F-360iSOD Dataset}}:
In Table~\ref{tab:table4}, we compare the proposed method with other state-of-the-art SOD methods on the F-360iSOD dataset. The results demonstrate that our model achieves optimal performance across all evaluation metrics, significantly outperforming existing methods in terms of S-measure, MAE, E-measure, and F-measure.

\textbf{Qualitative evaluation}:
\begin{figure}[t]
    \centering
    \includegraphics[width=\linewidth]{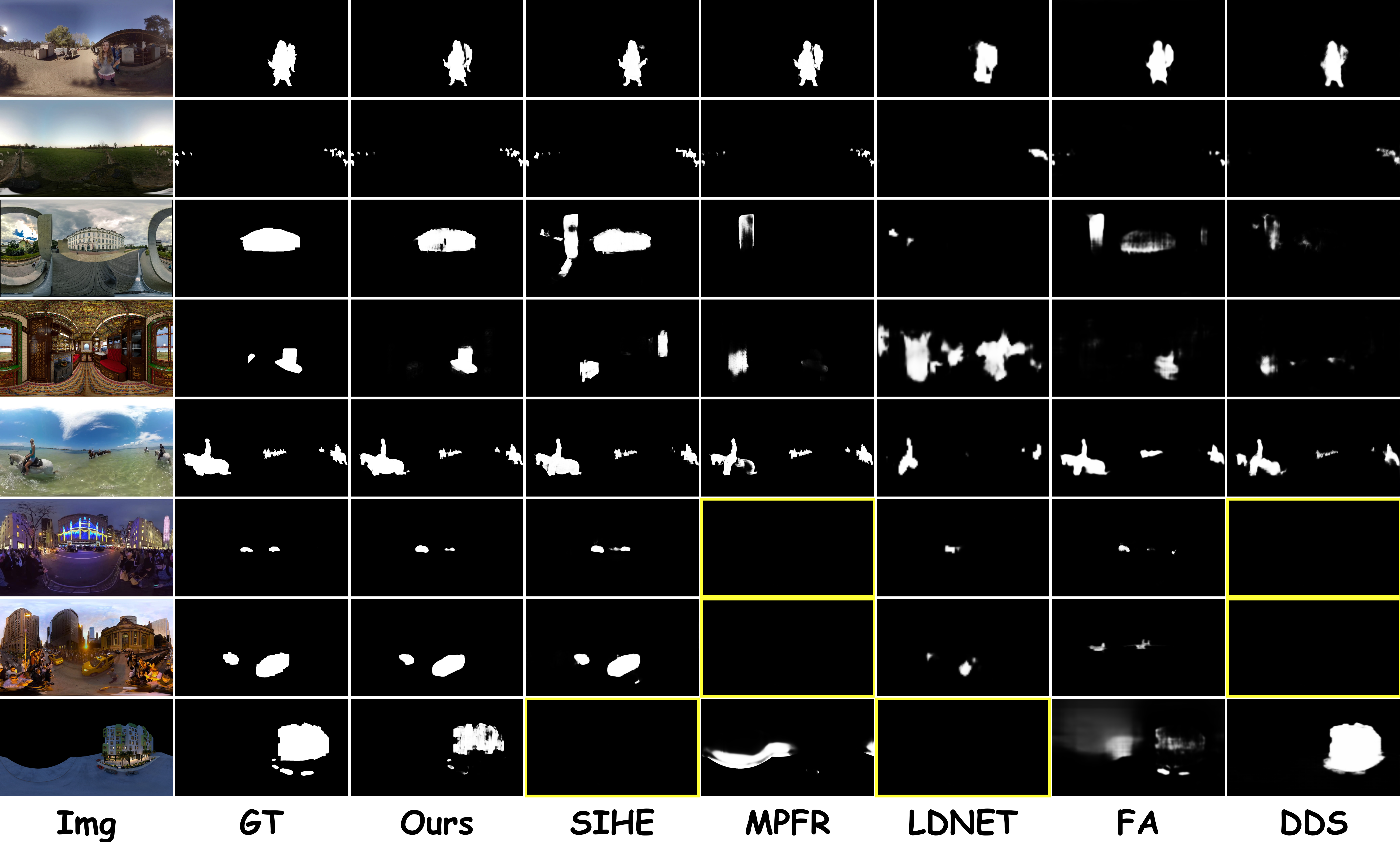}
    \caption{Qualitative comparisons of various PSOD methods under several challenging scenarios. The all-black images marked with yellow boxes indicate that the corresponding saliency maps cannot be obtained.}
    \Description{Qualitative comparisons}
    \label{fig:image5}
\end{figure}
To further evaluate practical detection performance, we present saliency maps predicted by different SOTA methods under representative challenging scenarios in Fig.~\ref{fig:image5}. For a convincing comparison, we select diverse scenes from multiple panoramic salient object detection datasets, covering multi object coexistence, complex background interference, low contrast conditions, and tiny objects. Some methods show clear limitations in these scenarios. For example, in rows 3 and 4, several methods produce saliency leakage under complex backgrounds and low contrast conditions, incorrectly classifying background regions as foreground. In row 8, other methods suffer from fragmented segmentation under complex background interference and fail to fully recover the contours of salient objects. In contrast, our model preserves clear boundaries for multiple salient instances, effectively suppresses responses from non target regions, and produces coherent and complete saliency maps even for low contrast and small objects. These qualitative results are consistent with the quantitative metrics and further validate the superiority of our method for panoramic salient object detection.

\subsection{Ablation study}
\label{sec:4.5}
\textbf{Effectiveness of the three-branch architecture: }To validate the effectiveness of the proposed ERP-CMP-Tangent three-branch architecture, we remove the Tangent branch and the CMP branch from the complete model, respectively, and construct three variants for comparative experiments: ERP single-branch, ERP+CMP dual-branch, and ERP-CMP-TAN three-branch, where both dual-branch and three-branch variants employ element-wise addition for feature fusion. As shown in Table~\ref{tab:table5}, the ERP single-branch achieves the lowest performance on both 360-SOD and 360-SSOD datasets, the ERP+CMP dual-branch obtains moderate improvements, while the complete three-branch architecture attains optimal results across all evaluation metrics. This can be attributed to the following factors: the ERP single-branch suffers from polar distortion, where features extracted by Hybrid-ViT exhibit geometric distortion and semantic ambiguity in the polar regions; with the introduction of CMP, cubic mapping alleviates polar distortion, yet the semantic discontinuity at cube face boundaries still constrains the discriminative capability of fused features; the Tangent branch extracts multi-view semantic consistency priors through 18 distortion-free local viewports, whose globally pooled features serve as cross-projection semantic calibration signals, effectively bridging the semantic gap caused by ERP polar distortion and CMP face boundary discontinuities, thereby enabling the three-branch architecture to achieve significant performance gains.
\begin{table}[t]
  \centering
  \caption{Ablation study on different branch combinations in the proposed model.}
  \label{tab:table5}
  \small
  \setlength{\tabcolsep}{3.1pt}
  \renewcommand{\arraystretch}{1.15}
  \begin{tabular}{c|c|cccc|c|cccc}
    \toprule
    \raisebox{-5.5ex}[0pt][0pt]{\rotatebox[origin=c]{90}{\textbf{360-SOD}}}
      & \textbf{Settings}
      & \textbf{$S_m\uparrow$} &\textbf{$MAE\downarrow$} 
      &\textbf{$meE\uparrow$} 
      & \textbf{$meF\uparrow$}
      & \raisebox{-5.7ex}[0pt][0pt]{\rotatebox[origin=c]{90}{\textbf{360-SSOD}}}
      & \textbf{$S_m\uparrow$} &\textbf{$MAE\downarrow$} 
      &\textbf{$meE\uparrow$} 
      & \textbf{$meF\uparrow$} \\
    \cline{2-6} \cline{8-11}
    & ERP         & .824 & .026 & .889 & .742 & & .748 & .039 & .827 & .595 \\
    & ERP-CMP     & .858 & .018 & .918 & .812 & & .793 & .028 & .862 & .677 \\
    
    & ERP-CMP-TAN
    &\textbf{.878} & \textbf{.016} & \textbf{.925} & \textbf{.822} & & \textbf{.796} & \textbf{.027} & \textbf{.872} & \textbf{.689} \\
    \bottomrule
  \end{tabular}
\end{table}

\begin{figure}[t]
    \centering
    \includegraphics[width=\linewidth]{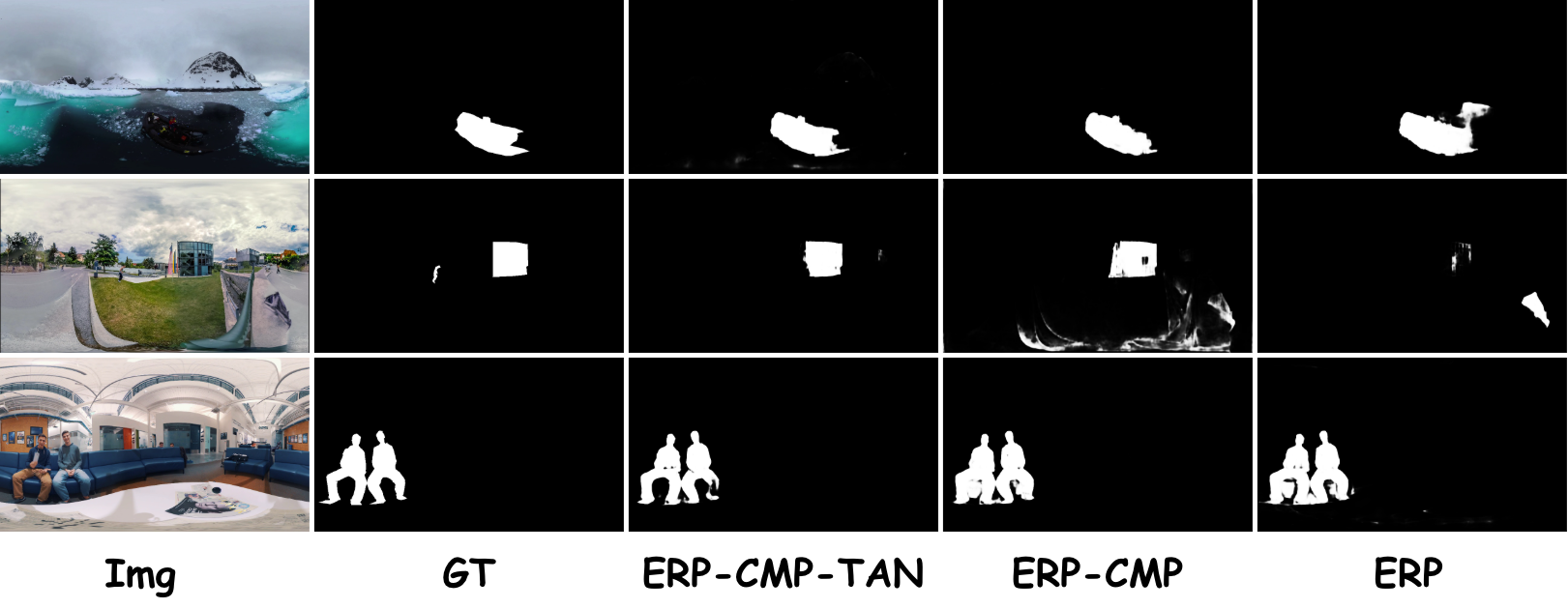}
    \caption{Visualization of three variants, ERP, ERP-CMP and ERP-CMP-TAN.}
    \Description{Visualization of three variants}
    \label{fig:image6}
\end{figure}
\begin{figure}[t]
    \centering
    \includegraphics[width=\linewidth]{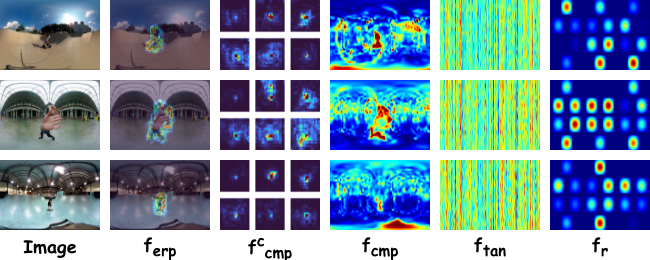}
    \caption{Visualization of multi-projection features extracted by the tri-branch encoder.}
    \Description{Visualization of different modules.}
    \label{fig:image7}
\end{figure}
Furthermore, Fig.~\ref{fig:image6} visualizes saliency maps predicted under different branch configurations. Compared with the complete three branch model, the ERP only branch produces noticeable false detections in complex backgrounds, as shown in row 2, and fails to preserve clear boundaries for objects with fine edges, as shown in row 3. The ERP and CMP branch alleviates polar distortion but still suffers from blurred edges and background response leakage, particularly in rows 1 and 2. This degradation mainly arises from two factors. First, the ERP only branch is affected by polar geometric stretching, which severely distorts features in high latitude regions and causes localization errors. Second, although the ERP and CMP branch introduces local geometric information, discontinuities along cube face boundaries introduce structural noise into the fused features. Without distortion free local semantic references from the Tangent branch for cross projection calibration, the model cannot fully eliminate edge fragmentation or false background responses.
Fig.~\ref{fig:image7} shows the intermediate representations extracted by the tri-branch encoder. ERP, CMP, and tangent-plane features exhibit clearly different response patterns due to their distinct projection geometries. ERP features preserve global spatial continuity, whereas CMP and tangent-plane features provide more localized and less distorted responses around salient regions. The reference feature further provides spatial cues for subsequent cross-projection alignment.

\begin{figure}[t]
    \centering
    \includegraphics[width=\linewidth]{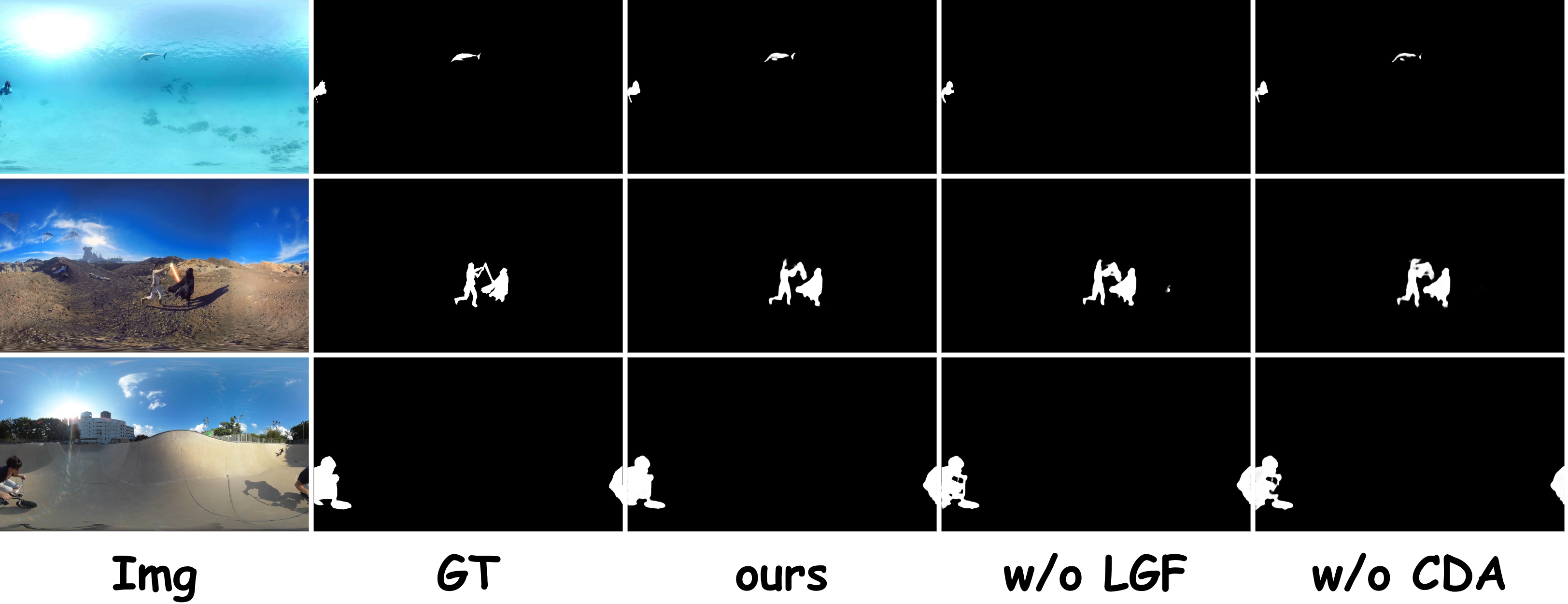}
    \caption{Visualization results of ablation studies in different modules.}
    \Description{Visualization of different modules.}
    \label{fig:image8}
\end{figure}
\begin{figure}[t]
    \centering
    \includegraphics[width=\linewidth]{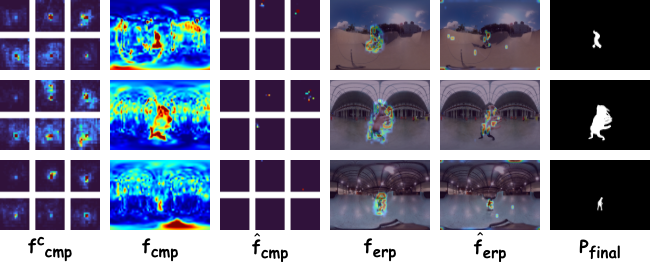}
    \caption{Visualization of feature enhancement by the proposed CDA module.}
    \label{fig:image9}
\end{figure}

\textbf{Effectiveness of individual modules in the framework:} To validate the effectiveness of the CDA and LGF modules in our network, we conduct ablation experiments by directly removing CDA from the encoding branches and replacing LGF with element-wise summation across the three branches. As shown in Table~\ref{tab:table6}, the model without CDA exhibits significant degradation across all evaluation metrics on both the 360-SOD and 360-SSOD datasets. Similarly, the model without LGF demonstrates inferior performance compared to the complete model, though it outperforms the configuration without CDA. In Fig.~\ref{fig:image8}, we present saliency map predictions under different module configurations. Compared to our complete model, the model without CDA produces noticeable edge blurring in scenes containing multiple objects and complex boundaries (e.g., rows 2 and 3), while the model without LGF suffers from missed detections under background interference and exhibits incomplete segmentation artifacts at object boundaries (e.g., rows 1 and 2). In contrast, our complete model, through the synergistic effect of distortion-aware enhancement from CDA and cross-projection adaptive fusion from LGF, accurately localizes salient objects while maintaining clear edge contours. The above experiments demonstrate the effectiveness of each module and the superiority of the proposed method.
Fig.~\ref{fig:image9} illustrates the effect of CDA on ERP and CMP representations. Compared with the original features, the enhanced features show more concentrated activations on salient objects and reduced responses in irrelevant background regions. This indicates that the shared cross-projection geometric prior helps adaptive sampling better align complementary information across projections and improves local geometric consistency.

\begin{table}[t]
  \centering
  \caption{Ablation experiment of each module in the model.}
  \label{tab:table6}
  \small
  \setlength{\tabcolsep}{3.1pt}
  \renewcommand{\arraystretch}{1.15}
  \begin{tabular}{c|c|cccc|c|cccc}
    \toprule
    \raisebox{-5.5ex}[0pt][0pt]{\rotatebox[origin=c]{90}{\textbf{360-SOD}}}
      & \textbf{Settings}
      & \textbf{$S_m\uparrow$} &\textbf{$MAE\downarrow$} 
      &\textbf{$meE\uparrow$} 
      & \textbf{$meF\uparrow$}
      & \raisebox{-5.7ex}[0pt][0pt]{\rotatebox[origin=c]{90}{\textbf{360-SSOD}}}
      & \textbf{$S_m\uparrow$} &\textbf{$MAE\downarrow$} 
      &\textbf{$meE\uparrow$} 
      & \textbf{$meF\uparrow$} \\
    \cline{2-6} \cline{8-11}
    & w/o CDA   & .869 & .017 & .923 & .807 & & .795 & .027 & .865 & .677 \\
    & w/o LGF    & .878 & .016 & .925 & .822 & & .796 & .027 & .872 & .689 \\
    & \textbf{Ours}        & \textbf{.883} & \textbf{.015} & \textbf{.927} & \textbf{.835} & & \textbf{.804} & \textbf{.026} & \textbf{.877} & \textbf{.697} \\
    \bottomrule
  \end{tabular}
\end{table}
\begin{table}[t]
  \centering
  \caption{Ablation study on the fusion strategies of the LGF module.}
  \label{tab:table7}
  \small
  \setlength{\tabcolsep}{4.5pt}
  \renewcommand{\arraystretch}{1.15}
  \begin{tabular}{c|cccc|cccc|cc}
    \toprule
    \multirow{2}{*}{\textbf{Strategy}} & \multicolumn{4}{c|}{\textbf{360-SOD}} & \multicolumn{4}{c|}{\textbf{360-SSOD}} & \multicolumn{2}{c}{\textbf{Complexity}} \\
    & \textbf{$S_m\uparrow$} &\textbf{$MAE\downarrow$} 
      &\textbf{$meE\uparrow$} 
      & \textbf{$meF\uparrow$} & \textbf{$S_m\uparrow$} &\textbf{$MAE\downarrow$} 
      &\textbf{$meE\uparrow$} 
      & \textbf{$meF\uparrow$}& Params (M) & FLOPs (G) \\
    \midrule
    Geometry Only & \textbf{0.883} & \textbf{0.015} & \textbf{0.927} & \textbf{0.835} & 0.789 & 0.029 & 0.863 & 0.678 & \textbf{0.000} & \textbf{$\sim$0} \\
    Mixed          & 0.860 & 0.019 & 0.917 & 0.802 & \textbf{0.804} & \textbf{0.026} & \textbf{0.877} & \textbf{0.697} & 0.084 & 0.086 \\
    \bottomrule
  \end{tabular}
\end{table}

\begin{figure}[t]
    \centering
    \includegraphics[width=\linewidth]{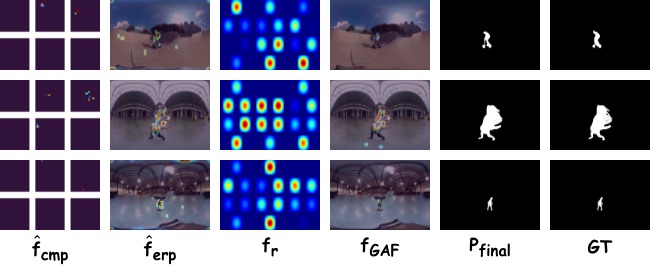}
    \caption{Visualization of the LGF fusion process and final predictions.}
    \label{fig:image10}
\end{figure}
\textbf{Effectiveness of LGF fusion strategies:}
To evaluate the Geometry Only and Mixed strategies in the LGF module, we conduct comparative experiments on the 360-SOD and 360-SSOD datasets. As shown in Table~\ref{tab:table7}, the Geometry Only strategy performs better on the smaller 360-SOD dataset, whereas the Mixed strategy achieves superior results on the larger 360-SSOD dataset. This difference can be attributed to their distinct fusion mechanisms. The Geometry Only strategy constructs parameter-free geometric confidence masks from spherical latitude maps and directly computes fusion weights using a fixed cosine function, enabling stable projection-adaptive balancing between ERP and CMP with limited training data. In contrast, the Mixed strategy introduces a lightweight learnable convolutional network for weight generation and learnable scalar coefficients for dynamically modulating tangent refinement features, providing greater flexibility for modeling complex cross-projection relationships on larger datasets. These results demonstrate the adaptability of LGF to different dataset scales. Notably, the Geometry Only strategy introduces no learnable parameters and negligible computational overhead, while the Mixed strategy adds only 0.084M parameters and 0.086G FLOPs, further demonstrating the lightweight design of LGF.

The visualization in Fig.~\ref{fig:image10} further illustrates the progressive multi-projection fusion process of LGF. The enhanced ERP and CMP features, together with the reference feature, provide complementary structural and semantic cues. After LGF fusion, the resulting feature $f_{\mathrm{GAF}}$ exhibits cleaner and more complete responses around salient objects while suppressing irrelevant background regions. Consequently, the final predictions show improved object completeness and closely match the ground-truth masks, further validating the effectiveness of the proposed fusion strategy.
\section{CONCLUSION}
In this paper, we present TDFNet, the first tri-projection deformable fusion network for panoramic salient object detection (PSOD). TDFNet exploits complementary information from two perspectives: cross projection feature complementarity and cross level feature reconstruction. Since ERP representations suffer from severe polar distortion and CMP projections exhibit boundary discontinuities between cube faces, we propose CDA to establish spatial mappings between ERP and CMP. These mappings are embedded into deformable attention sampling to aggregate contextual information across projections, thereby improving the encoder robustness to projection deformation. To address projection domain heterogeneity in three branch feature fusion, we propose LGF, which fuses spherical features from ERP and CMP and performs adaptive cross projection weighted refinement using distortion free semantic references from Tangent Projection. This design bridges the semantic gap between local perspectives and global representations. Because Tangent Projection provides distortion free local viewport information, we introduce it into panoramic salient object detection for the first time. It works collaboratively with the ERP and CMP branches to provide rich local and global complementary information for subsequent fusion. Extensive experiments on four representative PSOD benchmarks demonstrate that our method outperforms existing state of the art PSOD approaches.

\bibliographystyle{ACM-Reference-Format}
\bibliography{references}

@article{wu2022view,
  title={View-Aware Salient Object Detection for 360\(^\circ\) Omnidirectional Image},
  author={Wu, Junjie and Xia, Changqun and Yu, Tianshu and Li, Jia},
  journal={IEEE Transactions on Multimedia},
  volume={25},
  pages={6471--6484},
  year={2022},
  publisher={IEEE}
}

@inproceedings{zhao2023distortion,
  title={Distortion-aware transformer in 360\(^\circ\) salient object detection},
  author={Zhao, Yinjie and Zhao, Lichen and Yu, Qian and Sheng, Lu and Zhang, Jing and Xu, Dong},
  booktitle={Proceedings of the 31st ACM International Conference on Multimedia},
  pages={499--508},
  year={2023}
}

@inproceedings{zhang2022channel,
  title={Channel-spatial mutual attention network for 360 salient object detection},
  author={Zhang, Yi and Hamidouche, Wassim and Deforges, Olivier},
  booktitle={2022 26th International Conference on Pattern Recognition (ICPR)},
  pages={3436--3442},
  year={2022},
  organization={IEEE}
}

@inproceedings{zhang2023salient,
  title={Salient Object Detection on 360° Omnidirectional Image with Bi-Branch Hybrid Projection Network},
  author={Zhang, Jie and Zhang, Qiudan and Shen, Xuelin and Wang, Xu},
  booktitle={2023 IEEE 25th International Workshop on Multimedia Signal Processing (MMSP)},
  pages={1--5},
  year={2023},
  organization={IEEE}
}

@article{huang2020fanet,
  title={FANet: Features Adaptation Network for 360\(^\circ\) Omnidirectional Salient Object Detection},
  author={Huang, Mengke and Liu, Zhi and Li, Gongyang and Zhou, Xiaofei and Le Meur, Olivier},
  journal={IEEE Signal Processing Letters},
  volume={27},
  pages={1819--1823},
  year={2020},
  publisher={IEEE}
}

@article{wen2025consistency,
  title={Consistency perception network for 360\(^\circ\) omnidirectional salient object detection},
  author={Wen, Hongfa and Zhu, Zunjie and Zhou, Xiaofei and Zhang, Jiyong and Yan, Chenggang},
  journal={Neurocomputing},
  volume={620},
  pages={129243},
  year={2025},
  publisher={Elsevier}
}

@article{he2024sihenet,
  title={Sihenet: Semantic interaction and hierarchical embedding network for 360\(^\circ\) salient object detection},
  author={He, Zhentao and Shao, Feng and Xie, Zhengxuan and Chai, Xiongli and Ho, Yo-Sung},
  journal={IEEE Transactions on Instrumentation and Measurement},
  volume={74},
  pages={1--15},
  year={2024},
  publisher={IEEE}
}

@article{zhou2026differseg,
  title={DifferSeg: Towards Diverse Multimodal Binary Segmentation via Differential Perception and Frequency Guidance},
  author={Zhou, Qiangqiang and Xu, Jiawei and Chen, Yong and Zhu, Dandan and Yi, Yugen and Zhao, Xiaoqi},
  journal={IEEE Transactions on Circuits and Systems for Video Technology},
  year={2026},
  publisher={IEEE}
}

@article{hao2025simple,
  title={A simple yet effective network based on vision transformer for camouflaged object and salient object detection},
  author={Hao, Chao and Yu, Zitong and Liu, Xin and Xu, Jun and Yue, Huanjing and Yang, Jingyu},
  journal={IEEE Transactions on Image Processing},
  year={2025},
  publisher={IEEE}
}

@article{xu2025semantic,
  title={Semantic-Orthogonal Multi-modal Attention Network for RGB-D Salient Object Detection: J. Xu et al.},
  author={Xu, Jiawei and Zhou, Qiangqiang and Yu, Jiacong and Liao, Chen and Zhu, Dandan},
  journal={The Visual Computer},
  volume={41},
  number={9},
  pages={6917--6929},
  year={2025},
  publisher={Springer}
}

@article{tang2024divide,
  title={Divide-and-conquer: Confluent triple-flow network for RGB-T salient object detection},
  author={Tang, Hao and Li, Zechao and Zhang, Dong and He, Shengfeng and Tang, Jinhui},
  journal={IEEE transactions on pattern analysis and machine intelligence},
  volume={47},
  number={3},
  pages={1958--1974},
  year={2024},
  publisher={IEEE}
}

@article{xu2026hvpnet,
  title={HVPNet: A Bio-Inspired Network for General Salient and Camouflaged Object Detection},
  author={Xu, Jiawei and Zhou, Qiangqiang and Li, Zhouping and Shi, Yanjiao and Yi, Yugen and Yu, Jiacong},
  journal={Neural Networks},
  pages={109340},
  year={2026},
  publisher={Elsevier}
}

@inproceedings{xu2026tp,
  title={TP-Seg: Task-Prototype framework for unified medical lesion segmentation},
  author={Xu, Jiawei and Zhou, Qiangqiang and Zhu, Dandan and Chen, Yong and Yi, Yugen and Zhao, Xiaoqi},
  booktitle={Proceedings of the IEEE/CVF Conference on Computer Vision and Pattern Recognition},
  pages={5452--5462},
  year={2026}
}

@article{yamanaka2023multi,
  title={Multi-scale estimation for omni-directional saliency maps using learnable equator bias},
  author={Yamanaka, Takao and Suzuki, Tatsuya and Nobutsune, Taiki and Wu, Chenjunlin},
  journal={IEICE TRANSACTIONS on Information and Systems},
  volume={106},
  number={10},
  pages={1723--1731},
  year={2023},
  publisher={The Institute of Electronics, Information and Communication Engineers}
}

@article{zou2023360,
  title={360\(^\circ\) image saliency prediction by embedding self-supervised proxy task},
  author={Zou, Zizhuang and Ye, Mao and Li, Shuai and Li, Xue and Dufaux, Fr{\'e}deric},
  journal={IEEE Transactions on Broadcasting},
  volume={69},
  number={3},
  pages={704--714},
  year={2023},
  publisher={IEEE}
}

@article{zhang2024multi,
  title={Multi-scale graph feature extraction network for panoramic image saliency detection.},
  author={Zhang, Ripei and Chen, Chunyi and Peng, Jun},
  journal={Visual Computer},
  volume={40},
  number={2},
  pages={953},
  year={2024}
}

@article{li2019distortion,
  title={Distortion-Adaptive Salient Object Detection in 360\(^\circ\) Omnidirectional Images},
  author={Li, Jia and Su, Jinming and Xia, Changqun and Tian, Yonghong},
  journal={IEEE Journal of Selected Topics in Signal Processing},
  volume={14},
  number={1},
  pages={38--48},
  year={2019},
  publisher={IEEE}
}

@inproceedings{dai2023360,
  title={360\(^\circ\) Omnidirectional Salient Object Detection with Multi-scale Interaction and Densely-Connected Prediction},
  author={Dai, Haowei and Bao, Liuxin and Shen, Kunye and Zhou, Xiaofei and Zhang, Jiyong},
  booktitle={International Conference on Image and Graphics},
  pages={427--438},
  year={2023},
  organization={Springer}
}

@article{he2023scfanet,
  title={SCFANet: Semantics and context feature aggregation network for 360\(^\circ\) salient object detection},
  author={He, Zhentao and Shao, Feng and Chen, Gang and Chai, Xiongli and Ho, Yo-Sung},
  journal={IEEE Transactions on Multimedia},
  volume={26},
  pages={2276--2288},
  year={2023},
  publisher={IEEE}
}

@article{cokelek2025spherical,
  title={Spherical Vision Transformers for Audio-Visual Saliency Prediction in 360\(^\circ\) Videos},
  author={Cokelek, Mert and Ozsoy, Halit and Imamoglu, Nevrez and Ozcinar, Cagri and Ayhan, Inci and Erdem, Erkut and Erdem, Aykut},
  journal={IEEE Transactions on Pattern Analysis and Machine Intelligence},
  year={2025},
  publisher={IEEE}
}

@article{ma2020stage,
  title={Stage-wise salient object detection in 360 omnidirectional image via object-level semantical saliency ranking},
  author={Ma, Guangxiao and Li, Shuai and Chen, Chenglizhao and Hao, Aimin and Qin, Hong},
  journal={IEEE Transactions on Visualization and Computer Graphics},
  volume={26},
  number={12},
  pages={3535--3545},
  year={2020},
  publisher={IEEE}
}

@inproceedings{zhang2020fixation,
  title={A fixation-based 360 benchmark dataset for salient object detection},
  author={Zhang, Yi and Zhang, Lu and Hamidouche, Wassim and Deforges, Olivier},
  booktitle={2020 IEEE International Conference on Image Processing (ICIP)},
  pages={3458--3462},
  year={2020},
  organization={IEEE}
}

@article{kingma2014adam,
  title={Adam: A method for stochastic optimization},
  author={Kingma, Diederik P and Ba, Jimmy},
  journal={arXiv preprint arXiv:1412.6980},
  year={2014}
}

@inproceedings{fan2017structure,
  title={Structure-measure: A new way to evaluate foreground maps},
  author={Fan, Deng-Ping and Cheng, Ming-Ming and Liu, Yun and Li, Tao and Borji, Ali},
  booktitle={Proceedings of the IEEE international conference on computer vision},
  pages={4548--4557},
  year={2017}
}

@inproceedings{achanta2009frequency,
  title={Frequency-tuned salient region detection},
  author={Achanta, Radhakrishna and Hemami, Sheila and Estrada, Francisco and Susstrunk, Sabine},
  booktitle={2009 IEEE conference on computer vision and pattern recognition},
  pages={1597--1604},
  year={2009},
  organization={IEEE}
}

@article{fan2018enhanced,
  title={Enhanced-alignment measure for binary foreground map evaluation},
  author={Fan, Deng-Ping and Gong, Cheng and Cao, Yang and Ren, Bo and Cheng, Ming-Ming and Borji, Ali},
  journal={arXiv preprint arXiv:1805.10421},
  year={2018}
}

@article{cong2023multi,
  title={Multi-projection fusion and refinement network for salient object detection in 360° omnidirectional image},
  author={Cong, Runmin and Huang, Ke and Lei, Jianjun and Zhao, Yao and Huang, Qingming and Kwong, Sam},
  journal={IEEE Transactions on Neural Networks and Learning Systems},
  volume={35},
  number={7},
  pages={9495--9507},
  year={2023},
  publisher={IEEE}
}

@article{huang2023lightweight,
  title={Lightweight distortion-aware network for salient object detection in omnidirectional images},
  author={Huang, Mengke and Li, Gongyang and Liu, Zhi and Zhu, Linchao},
  journal={IEEE Transactions on Circuits and Systems for Video Technology},
  volume={33},
  number={10},
  pages={6191--6197},
  year={2023},
  publisher={IEEE}
}

@article{yao2021boundary,
  title={Boundary information progressive guidance network for salient object detection},
  author={Yao, Zhaojian and Wang, Luping},
  journal={IEEE Transactions on Multimedia},
  volume={24},
  pages={4236--4249},
  year={2021},
  publisher={IEEE}
}

@inproceedings{xie2022pyramid,
  title={Pyramid grafting network for one-stage high resolution saliency detection},
  author={Xie, Chenxi and Xia, Changqun and Ma, Mingcan and Zhao, Zhirui and Chen, Xiaowu and Li, Jia},
  booktitle={Proceedings of the IEEE/CVF conference on computer vision and pattern recognition},
  pages={11717--11726},
  year={2022}
}

@article{zhu2021supplement,
  title={Supplement and suppression: Both boundary and nonboundary are helpful for salient object detection},
  author={Zhu, Ge and Li, Jinbao and Guo, Yahong},
  journal={IEEE Transactions on Neural Networks and Learning Systems},
  volume={34},
  number={9},
  pages={6615--6627},
  year={2021},
  publisher={IEEE}
}

@inproceedings{yu2021structure,
  title={Structure-consistent weakly supervised salient object detection with local saliency coherence},
  author={Yu, Siyue and Zhang, Bingfeng and Xiao, Jimin and Lim, Eng Gee},
  booktitle={Proceedings of the AAAI conference on artificial intelligence},
  volume={35},
  number={4},
  pages={3234--3242},
  year={2021}
}

@article{liu2020lightweight,
  title={Lightweight salient object detection via hierarchical visual perception learning},
  author={Liu, Yun and Gu, Yu-Chao and Zhang, Xin-Yu and Wang, Weiwei and Cheng, Ming-Ming},
  journal={IEEE transactions on cybernetics},
  volume={51},
  number={9},
  pages={4439--4449},
  year={2020},
  publisher={IEEE}
}

@article{li2022adjacent,
  title={Adjacent context coordination network for salient object detection in optical remote sensing images},
  author={Li, Gongyang and Liu, Zhi and Zeng, Dan and Lin, Weisi and Ling, Haibin},
  journal={IEEE Transactions on Cybernetics},
  volume={53},
  number={1},
  pages={526--538},
  year={2022},
  publisher={IEEE}
}

@article{qin2021boundary,
  title={Boundary-aware segmentation network for mobile and web applications},
  author={Qin, Xuebin and Fan, Deng-Ping and Huang, Chenyang and Diagne, Cyril and Zhang, Zichen and Sant'Anna, Adri{\`a} Cabeza and Suarez, Albert and Jagersand, Martin and Shao, Ling},
  journal={arXiv preprint arXiv:2101.04704},
  year={2021}
}

@article{zhu2023dc,
  title={Dc-net: Divide-and-conquer for salient object detection},
  author={Zhu, Jiayi and Qin, Xuebin and Elsaddik, Abdulmotaleb},
  journal={arXiv preprint arXiv:2305.14955},
  year={2023}
}

@article{ma2023boosting,
  title={Boosting broader receptive fields for salient object detection},
  author={Ma, Mingcan and Xia, Changqun and Xie, Chenxi and Chen, Xiaowu and Li, Jia},
  journal={IEEE Transactions on Image Processing},
  volume={32},
  pages={1026--1038},
  year={2023},
  publisher={IEEE}
}

@inproceedings{gao2024multi,
  title={Multi-scale and detail-enhanced segment anything model for salient object detection},
  author={Gao, Shixuan and Zhang, Pingping and Yan, Tianyu and Lu, Huchuan},
  booktitle={Proceedings of the 32nd ACM international conference on multimedia},
  pages={9894--9903},
  year={2024}
}

@inproceedings{ni2025makes,
  title={What makes for text to 360-degree panorama generation with stable diffusion?},
  author={Ni, Jinhong and Zhang, Chang-Bin and Zhang, Qiang and Zhang, Jing},
  booktitle={2025 IEEE/CVF International Conference on Computer Vision (ICCV)},
  pages={1--10},
  year={2025},
  organization={IEEE}
}

@inproceedings{yu2023applications,
  title={Applications of deep learning for top-view omnidirectional imaging: A survey},
  author={Yu, Jingrui and Grassi, Ana Cecilia Perez and Hirtz, Gangolf},
  booktitle={2023 IEEE/CVF Conference on Computer Vision and Pattern Recognition Workshops (CVPRW)},
  pages={6421--6433},
  year={2023},
  organization={IEEE}
}

@inproceedings{bao2025free360,
  title={Free360: Layered gaussian splatting for unbounded 360-degree view synthesis from extremely sparse and unposed views},
  author={Bao, Chong and Zhang, Xiyu and Yu, Zehao and Shi, Jiale and Zhang, Guofeng and Peng, Songyou and Cui, Zhaopeng},
  booktitle={2025 IEEE/CVF Conference on Computer Vision and Pattern Recognition (CVPR)},
  pages={16377--16387},
  year={2025},
  organization={IEEE}
}

@article{wang2020attention,
  title={Attention-Based Deep Reinforcement Learning for Virtual Cinematography of 360$^{\circ}$ Videos},
  author={Wang, Jianyi and Xu, Mai and Jiang, Lai and Song, Yuhang},
  journal={IEEE Transactions on Multimedia},
  volume={23},
  pages={3227--3238},
  year={2020},
  publisher={IEEE}
}

@inproceedings{gungordu2024saliency,
  title={Saliency-aware end-to-end learned variable-bitrate 360-degree image compression},
  author={G{\"u}ng{\"o}rd{\"u}, O{\u{g}}uzhan and Tekalp, A Murat},
  booktitle={2024 IEEE International Conference on Image Processing (ICIP)},
  pages={1795--1801},
  year={2024},
  organization={IEEE}
}

@article{yan2025omnidirectional,
  title={Omnidirectional image quality captioning: A large-scale database and a new model},
  author={Yan, Jiebin and Tan, Ziwen and Fang, Yuming and Chen, Junjie and Jiang, Wenhui and Wang, Zhou},
  journal={IEEE Transactions on Image Processing},
  volume={34},
  pages={1326--1339},
  year={2025},
  publisher={IEEE}
}

@inproceedings{jiang2013salient,
  title={Salient object detection: A discriminative regional feature integration approach},
  author={Jiang, Huaizu and Wang, Jingdong and Yuan, Zejian and Wu, Yang and Zheng, Nanning and Li, Shipeng},
  booktitle={Proceedings of the IEEE conference on computer vision and pattern recognition},
  pages={2083--2090},
  year={2013}
}

@inproceedings{mazumdar2019content,
  title={A content-based approach for saliency estimation in 360 images},
  author={Mazumdar, Pramit and Battisti, Federica},
  booktitle={2019 IEEE International Conference on Image Processing (ICIP)},
  pages={3197--3201},
  year={2019},
  organization={IEEE}
}

@article{lebreton2018gbvs360,
  title={GBVS360, BMS360, ProSal: Extending existing saliency prediction models from 2D to omnidirectional images},
  author={Lebreton, Pierre and Raake, Alexander},
  journal={Signal Processing: Image Communication},
  volume={69},
  pages={69--78},
  year={2018},
  publisher={Elsevier}
}

@article{wei2025rdfcnet,
  title={RDFCNet: RGB-guided depth feature calibration network for RGB-D Salient Object Detection},
  author={Wei, Weiyi and Shi, Yongxin and Wang, Yibin and Liu, Sixuan},
  journal={Neurocomputing},
  pages={131127},
  year={2025},
  publisher={Elsevier}
}

@article{zhu2025scandtm,
  title={ScanDTM: A novel dual-temporal modulation scanpath prediction model for omnidirectional images},
  author={Zhu, Dandan and Zhang, Kaiwei and Min, Xiongkuo and Zhai, Guangtao and Yang, Xiaokang},
  journal={IEEE Transactions on Circuits and Systems for Video Technology},
  volume={35},
  number={8},
  pages={7850--7865},
  year={2025},
  publisher={IEEE}
}

@inproceedings{ronneberger2015u,
  title={U-net: Convolutional networks for biomedical image segmentation},
  author={Ronneberger, Olaf and Fischer, Philipp and Brox, Thomas},
  booktitle={International Conference on Medical image computing and computer-assisted intervention},
  pages={234--241},
  year={2015},
  organization={Springer}
}

@article{zhang2023360,
  title={360-degree visual saliency detection based on fast-mapped convolution and adaptive equator-bias perception},
  author={Zhang, Ripei and Chen, Chunyi and Zhang, Jiacheng and Peng, Jun and Alzbier, Ahmed Mustafa Taha},
  journal={The Visual Computer},
  volume={39},
  number={3},
  pages={1163--1180},
  year={2023},
  publisher={Springer}
}

@article{qing2022attentive,
  title={Attentive and context-aware deep network for saliency prediction on omni-directional images},
  author={Qing, Chunmei and Zhu, Huansheng and Xing, Xiaofen and Chen, Dongwen and Jin, Jianxiu},
  journal={Digital Signal Processing},
  volume={120},
  pages={103289},
  year={2022},
  publisher={Elsevier}
}

@article{chao2020multi,
  title={A Multi-FoV Viewport-Based Visual Saliency Model Using Adaptive Weighting Losses for 360$^{\circ}$ Images},
  author={Chao, Fang-Yi and Zhang, Lu and Hamidouche, Wassim and D{\'e}forges, Olivier},
  journal={IEEE Transactions on Multimedia},
  volume={23},
  pages={1811--1826},
  year={2020},
  publisher={IEEE}
}

@article{xu2021saliency,
  title={Saliency prediction on omnidirectional image with generative adversarial imitation learning},
  author={Xu, Mai and Yang, Li and Tao, Xiaoming and Duan, Yiping and Wang, Zulin},
  journal={IEEE Transactions on Image Processing},
  volume={30},
  pages={2087--2102},
  year={2021},
  publisher={IEEE}
}

@article{gao2019res2net,
  title={Res2net: A new multi-scale backbone architecture},
  author={Gao, Shang-Hua and Cheng, Ming-Ming and Zhao, Kai and Zhang, Xin-Yu and Yang, Ming-Hsuan and Torr, Philip},
  journal={IEEE transactions on pattern analysis and machine intelligence},
  volume={43},
  number={2},
  pages={652--662},
  year={2019},
  publisher={IEEE}
}

@inproceedings{lin2017feature,
  title={Feature pyramid networks for object detection},
  author={Lin, Tsung-Yi and Doll{\'a}r, Piotr and Girshick, Ross and He, Kaiming and Hariharan, Bharath and Belongie, Serge},
  booktitle={2017 IEEE conference on computer vision and pattern recognition (CVPR)},
  pages={936--944},
  year={2017},
  organization={Ieee}
}

@article{howard2017mobilenets,
  title={Mobilenets: Efficient convolutional neural networks for mobile vision applications},
  author={Howard, Andrew G and Zhu, Menglong and Chen, Bo and Kalenichenko, Dmitry and Wang, Weijun and Weyand, Tobias and Andreetto, Marco and Adam, Hartwig},
  journal={arXiv preprint arXiv:1704.04861},
  year={2017}
}

@article{li2021salient,
  title={Salient object detection with purificatory mechanism and structural similarity loss},
  author={Li, Jia and Su, Jinming and Xia, Changqun and Ma, Mingcan and Tian, Yonghong},
  journal={IEEE Transactions on Image Processing},
  volume={30},
  pages={6855--6868},
  year={2021},
  publisher={IEEE}
}

@inproceedings{perazzi2012saliency,
  title={Saliency filters: Contrast based filtering for salient region detection},
  author={Perazzi, Federico and Kr{\"a}henb{\"u}hl, Philipp and Pritch, Yael and Hornung, Alexander},
  booktitle={2012 IEEE conference on computer vision and pattern recognition},
  pages={733--740},
  year={2012},
  organization={IEEE}
}

@article{lyu2026hallu_sae,
  title={Towards Interpretable Hallucination Analysis and Mitigation in LVLMs via Contrastive Neuron Steering},
  author={Lyu, Guangtao and Cheng, Xinyi and Liu, Qi and Xu, Chenghao and Yan, Jiexi and Yang, Muli and Fang, Fen and Deng, Cheng},
  journal={arXiv preprint arXiv:2602.00621},
  year={2026}
}

@article{lyu2025hallu_vdc,
  title={Revealing Perception and Generation Dynamics in LVLMs: Mitigating Hallucinations via Validated Dominance Correction},
  author={Lyu, Guangtao and Cheng, Xinyi and Xu, Chenghao and Liu, Qi and Yang, Muli and Fang, Fen and Chen, Huilin and Yan, Jiexi and Yang, Xu and Deng, Cheng},
  journal={arXiv preprint arXiv:2512.18813},
  year={2025}
}

@inproceedings{xu2026psgait,
  title={Psgait: Gait recognition using parsing skeleton},
  author={Xu, Hangrui and Wu, Zhengxian and Zhang, Chuanrui and Chen, Zhuohong and Liu, Zhifang and Jiao, Peng and Wang, Haoqian},
  booktitle={ICASSP 2026-2026 IEEE International Conference on Acoustics, Speech and Signal Processing (ICASSP)},
  pages={10427--10431},
  year={2026},
  organization={IEEE}
}

@inproceedings{zhong2025adaptive,
  title={Adaptive Prompt Learning for Blind Image Quality Assessment with Multi-modal Mixed-datasets Training},
  author={Zhong, Yan and Zhao, Xinping and Zhang, Li and Song, Xinyuan and Jiang, Tingting},
  booktitle={Proceedings of the 33rd ACM International Conference on Multimedia},
  pages={7453--7462},
  year={2025}
}

@inproceedings{zhong2024causal,
  title={Causal-IQA: Towards the Generalization of Image Quality Assessment Based on Causal Inference.},
  author={Zhong, Yan and Wu, Xingyu and Zhang, Li and Yang, Chenxi and Jiang, Tingting},
  booktitle={ICML},
  year={2024}
}

@article{zhong2025ctd,
  title={Ctd-inpainting: Towards the coherence of text-driven inpainting with blended diffusion},
  author={Zhong, Yan and Zhao, Xinping and Zhao, Guangzhi and Chen, Bohua and Hao, Fei and Zhao, Ruoyu and He, Jiaqi and Shi, Lei and Zhang, Li},
  journal={Information Fusion},
  volume={122},
  pages={103163},
  year={2025},
  publisher={Elsevier}
}

@inproceedings{long2025revisiting,
  title={Revisiting multimodal fusion for 3D anomaly detection from an architectural perspective},
  author={Long, Kaifang and Xie, Guoyang and Ma, Lianbo and Liu, Jiaqi and Lu, Zhichao},
  booktitle={Proceedings of the AAAI Conference on Artificial Intelligence},
  volume={39},
  number={12},
  pages={12273--12281},
  year={2025}
}

@inproceedings{long2026towards,
  title={Towards an Incremental Unified Multimodal Anomaly Detection: Augmenting Multimodal Denoising From an Information Bottleneck Perspective},
  author={Long, Kaifang and Ma, Lianbo and Liu, Jiaqi and Liu, Liming and Xie, Guoyang},
  booktitle={Proceedings of the IEEE/CVF Conference on Computer Vision and Pattern Recognition},
  pages={14116--14125},
  year={2026}
}

@article{long2025enhancing,
  title={Enhancing Multimodal Learning via Hierarchical Fusion Architecture Search With Inconsistency Mitigation},
  author={Long, Kaifang and Xie, Guoyang and Ma, Lianbo and Li, Qing and Huang, Min and Lv, Jianhui and Lu, Zhichao},
  journal={IEEE Transactions on Image Processing},
  year={2025},
}

@article{xiao2026staying,
  title={Staying VIGILant: Mitigating Visual Laziness via Counterfactual Visual Alignment in MLLMs},
  author={Xiao, Xi and Liu, Chen and Liao, Chih-Ting and Zhang, Yunbei and Lan, Qizhen and Wei, Yuxiang and Zhao, Lin and Wang, Janet and Gu, Jianyang and Ye, Muchao and others},
  journal={arXiv preprint arXiv:2606.26387},
  year={2026}
}

@article{xiao2026layer,
  title={Layer-Specific Prompt Fusion Discovery via Differentiable Search in Vision Foundation Models},
  author={Xiao, Xi and Li, Xingjian and Zhang, Yunbei and Han, Cheng and Liu, Tianming and Wang, Tianyang and Jiang, Runmin and Hamm, Jihun and Wang, Xiao and Xu, Min},
  journal={arXiv preprint arXiv:2606.26379},
  year={2026}
}

@inproceedings{wu2025dagait,
  title={DAGait: Generalized skeleton-guided data alignment for gait recognition},
  author={Wu, Zhengxian and Zhang, Chuanrui and Xu, Hangrui and Jiao, Peng and Wang, Haoqian},
  booktitle={2025 IEEE International Conference on Multimedia and Expo (ICME)},
  pages={1--6},
  year={2025},
  organization={IEEE}
}

@inproceedings{zhong2025semi,
  title={Semi-supervised blind quality assessment with confidence-quantifiable pseudo-label learning for authentic images},
  author={Zhong, Yan and Yang, Chenxi and Zhao, Suyuan and Jiang, Tingting},
  booktitle={Forty-second International Conference on Machine Learning},
  year={2025}
}

\end{document}